%% file: main_ieee.tex
\documentclass[conference]{IEEEtran}
\IEEEoverridecommandlockouts
\usepackage{cite}
\usepackage{amsmath,amssymb,amsfonts}
\usepackage{algorithmic}
\usepackage{graphicx}
\usepackage{textcomp}
\usepackage{xcolor}

\usepackage{microtype}
\usepackage{graphicx}
\usepackage{subfigure}
\usepackage{booktabs} % for professional tables

\usepackage{multirow}
\usepackage{amsmath}
\usepackage{tikz-cd}
\usepackage{mathtools}

\usepackage[draft]{hyperref}

\usepackage{commands}
\usepackage{cfvae_results}
\usepackage{afterpage}
\usepackage{tabularx}
\usepackage{adjustbox}
\usepackage{caption}
\usepackage{svg}

\def\BibTeX{{\rm B\kern-.05em{\sc i\kern-.025em b}\kern-.08em
    T\kern-.1667em\lower.7ex\hbox{E}\kern-.125emX}}

\begin{document}
\addtolength{\topmargin}{+0.1cm}
    \newcolumntype{R}{>{\raggedleft\arraybackslash}X}%
    \newcolumntype{L}{>{\raggedright\arraybackslash}X}%

    \title{
        {\ModelFullName}: A density estimation model for Non-Intrusive Load Monitoring%\\
        \thanks{
            This material is based upon work supported by Air Force Office Scientific Research under award number FA9550-19-1-0020.
            % Also, this study was financed in part by the Coordenação de Aperfeiçoamento de Pessoal de Nível Superior – Brasil (CAPES) – Finance Code 001.
        }
    }
    
    \author{
        \IEEEauthorblockN{1\textsuperscript{st} Luis Felipe M.O. Henriques}
        \IEEEauthorblockA{\textit{Department of Computer Science} \\
        \textit{Pontifical Catholic University of Rio de Janeiro}\\
        Rio de Janeiro, Brazil \\
        lhenriques@inf.puc-rio.br}
        \and
        \IEEEauthorblockN{2\textsuperscript{nd} Eduardo Morgan}
        \IEEEauthorblockA{\textit{Department of Computer Science} \\
        \textit{Pontifical Catholic University of Rio de Janeiro}\\
        Rio de Janeiro, Brazil \\
        emorgan@inf.puc-rio.br}
        \and
        \IEEEauthorblockN{3\textsuperscript{rd} Sergio Colcher}
        \IEEEauthorblockA{\textit{Department of Computer Science} \\
        \textit{Pontifical Catholic University of Rio de Janeiro}\\
        Rio de Janeiro, Brazil \\
        colcher@inf.puc-rio.br}
        \and
        \IEEEauthorblockN{4\textsuperscript{th} Ruy Luiz Milidiú}
        \IEEEauthorblockA{\textit{Department of Computer Science} \\
        \textit{Pontifical Catholic University of Rio de Janeiro}\\
        Rio de Janeiro, Brazil \\
        milidiu@inf.puc-rio.br}
    }
    
    \maketitle
    
    \input{content/abstract}
    
    
    \input{content/introduction/intro}
    \input{content/related_work}

    \input{content/background/background}

    \input{content/cfvae/cfvae}
    \input{content/architecture-details/architecture-details}
    \input{content/experiments/experiments}

    \input{content/conclusion}

    \bibliography{./references}
    \bibliographystyle{./IEEEtran}
    % \bibliographystyle{./bibliography/IEEEtran}
    % \bibliography{./bibliography/IEEEabrv,./bibliography/IEEEexample}
    
    \vspace{12pt}
    \include{supplementary_material/main_supplementary}

\end{document}

%% file: content/abstract.tex
\begin{abstract}
    
    Non-Intrusive Load Monitoring (NILM) is a computational technique
        that aims at estimating each separate appliance's power load consumption
        based on the whole consumption measured by a single meter.
    This paper proposes a conditional density estimation model
        that joins a Conditional Variational Autoencoder 
        with a Conditional Invertible Normalizing Flow model
        to estimate the individual appliance's electricity consumption.
    The resulting model is called {\ModelFullName} or, 
        for simplicity, 
        {\ModelName}.
    Unlike most NILM models, 
        the resulting model estimates all appliances' power demand,
        appliance-by-appliance, 
        at once, 
        instead of having one model per appliance, which makes it easier.
    We train and evaluate our proposed model in a publicly available dataset 
        composed of energy consumption measurements
        from a poultry feed factory located in Brazil.
    The proposed model achieves highly competitive results when comparing to previous work.
    For six of the eight machines in the dataset, 
        we observe consistent improvements 
        that range from {\PINDEImprovement}
        up to {\MINDEImprovement} in the Normalized Disaggregation Error (NDE)
        and {\PIISAEImprovement}
        up to {\DPCIISAEImprovement} in the Signal Aggregate Error (SAE).
\end{abstract}

%% file: content/introduction/intro.tex
\section{Introduction} \label{sec:intro}
    
    Non-intrusive load monitoring (NILM) 
        is a computational technique for estimating individual appliances'
        electricity consumption from a single metering-point 
        that measures the whole circuit's consumption.
    From a signal processing perspective, 
        each appliance %-- an electrical load -- 
        is the signal source of an electrical consumption 
        that constitutes part of the 
        aggregated signal -- the whole circuit's consumption.
    Several applications have been proposed for an accurate NILM algorithm \cite{Kelly2015NeuralND, Martins2018ApplicationOA}.
    % Thus, 
        % NILM can be treated as a blind source separation process  \cite{Martins2018ApplicationOA}.
    % From a signal processing perspective, 
    %     NILM can be treated as a process of blind source separation \cite{Martins2018ApplicationOA}.

    % Among its various applications, 
    %     an accurate NILM algorithm could help operators manage the grid, 
    %     identify faulty appliances, 
    %     survey appliance usage behavior, 
    %     or produce itemized electricity bills from a single meter \cite{Kelly2015NeuralND}. 
    % Moreover,
    %     real-time monitoring and feedback of machinery power consumption can give experts insightful information about the waste of energy, 
    %     thus helping leverage the demand response\cite{Martins2018ApplicationOA}.   
    Traditional work on NILM rely on hand-engineered features 
        on top of a small neural network 
        or a linear model 
        \cite{Roos1994UsingNN, Yang&Chang2007, Lin&Tsai2010, Ruzzelli&Nicolas&Schoofs2010, Chang&Chien&Lin&Chen2011},
        and some work focus on extracting only transitions between steady-states \cite{Hart1992NonintrusiveAL}.
    
    This work follows the approach found in \cite{Kelly2015NeuralND, Martins2018ApplicationOA}
        and proposes a Deep Neural Network to learn a NILM model from the raw signal.
    Our proposed architecture is a Conditional Density Estimation (CDE) model, 
        which joins a Conditional Variational Autoencoder (CVAEs) \cite{Kingma2013AutoEncodingVB, Sohn2015LearningSO} 
        with a Conditional Invertible Normalizing Flow model (CNF)
        \cite{Dinh2014NICENI, Dinh2016DensityEU, Kingma2018GlowGF}.
    We perform the experiments in a publicly available dataset \cite{Martins2018Dataset} that contains data collected from a Brazilian poultry feed factory.
    % We compare the achieved results 
    %     against the previous work \cite{Martins2018ApplicationOA} in this dataset. 
    Our contributions are two-fold:
    \begin{enumerate}
        \item 
            We propose the {\ModelName} architecture, 
                which joins a CVAE model with a CNF model in a new way, 
                and thus conditioning the generative process over the input data.
        
        \item
            We advance the results obtained in the dataset, 
                achieving highly competitive results against previous work 
                with a model capable of disaggregating multiple appliances at once, 
                instead of having a single model per appliance,
                such as what is usually done.
    \end{enumerate}
    
    \input{content/introduction/parper_structure}%

%% file: content/introduction/parper_structure.tex
    The remainder of this paper is organized as follows. 
    Section \ref{sec:related-work} presents related work on NILM.
        Section \ref{sec:background}
        introduces related models such as CVAEs
        and CNFs. 
        % and their extensions to the conditional case.  
        Section \ref{sec:pfvae} describes our proposed {\ModelName} model and its formulation.
    The dataset description, experiments and results are presented in Section \ref{sec:results} and a brief conclusion in Section \ref{sec:conclusion}.%
    % Section \ref{sec:results} offers a brief description of the dataset used, the experiments performed and its results. Finally, we present the conclusions in \ref{sec:conclusion}
%

%% file: content/related_work.tex
\section{Related Work} \label{sec:related-work}
    Introduced by Hart in 1992 \cite{Hart1992NonintrusiveAL}, 
        NILM aims at estimating individual appliance energy consumption from Low-Voltage Distribution Board data. 
    Hart models each appliance as a finite-state machine where an event detection model infers the steady-state transitions.
    This model is based on signature taxonomy features. In 1994,
        Roos et al. \cite{Roos1994UsingNN} proposed a shallow Neural Network (NN) based on hand-engineered features to address the NILM task.
    Since then, 
       much work has been done on engineering features as inputs to shallow NNs \cite{Yang&Chang2007, Lin&Tsai2010, Ruzzelli&Nicolas&Schoofs2010, Chang&Chien&Lin&Chen2011}. 
    
    More recently,
        in 2011, 
        Zeifman et al. \cite{Zeifman2011NonintrusiveAL} enumerated several applications for which appliance-specific consumption information is an important feature.
    % Among the applications mentioned are
    %     fault detection,
    %     behavioral pattern elucidation,
    %     appliance analysis based on use,
    %     energy-aware appliance redesign,
    %     load forecasting,
    %     economic models,
    %     and energy efficiency programs.
    % Following these ideas,
    Posteriorly,
        Armel et al. \cite{Armel2013IsDT} suggested that proper feedback 
        and detailed information can provide up to an $18\%$ reduction in electricity consumption for commercial and residential buildings.
    
    Kelly and Knottenbelt \cite{Kelly2015NeuralND} 
        avoided hand engineering features by adapting deep NN architectures 
        to NILM.
    % Their work compared the performance of 
    %     long short-term memory (LSTM) networks \cite{Hochreiter1997LongSM},
    %     denoising autoencoders \cite{Kingma2013AutoEncodingVB}, and
    %     a proposed NN architecture 
    %     against classical methods such as 
    %     a Factorial Hidden Markov Model (FHMM) \cite{Ghahramani1997FactorialHM}.
    The experiments were performed on the UK-Dale dataset \cite{UK-DALE}, 
        which contains real aggregated and appliance-level data
        from five appliances in 6 households in the UK. 
    % They trained their models in data from a set of houses and always left out one of the houses for testing. 
    % Following this methodology, 
    %     their three proposed deep learning models present better results when applied to data from the house that was not given to the model during the training.  
    
    Bonfigli, R. et al. \cite{Bonfigli2018DenoisingAF} presented a denoising autoencoder \cite{Kingma2013AutoEncodingVB} architecture for NILM, 
        introducing several improvements to the traditional scheme for deep NNs.
    They conducted experiments on three publicly available datasets 
        and compared their results against the AFAMAP algorithm \cite{Kolter2012ApproximateII}, 
        achieving better results on average.
    In the same year,
        Zhang, Chaoyun et al. \cite{Zhang2018SequencetopointLW} proposed a convolutional NN for sequence-to-point learning. 
    % Their model receives a window of the aggregated data as input and outputs a single point of the target appliance.
    They applied their proposed approach to real-world household energy data 
        and achieved state-of-the-art performance.
    
    Still in 2018,
        Martins et al. \cite{Martins2018ApplicationOA} 
            presented an industrial electric energy consumption dataset collected from a Brazilian poultry feed factory.  
    Using this dataset,
        they compared a FHMM \cite{Ghahramani1997FactorialHM}
        and a Wavenet model \cite{Oord2016WaveNetAG} adaptation to disaggregate four pairs of different industrial appliances.
    % In comparison to the FHMM, 
    %     their model reduced the normalized disaggregation error (NDE) 
    %     and the signal aggregated error (SAE) for the same appliances. 
    % Additionally, 
    Their model increased the time percentage for which the machine is correctly classified as turned ON or OFF.

    This work's approach is based on modeling NILM to estimate a complex noise distribution conditioned on the observed aggregated appliances. 
    Unlike most of the related work, 
        our approach is to train a single model to disaggregate all the individual appliances at once.
    This approach does not ignore that we may have multiple activated machines simultaneously in the whole circuit,
        and therefore, 
        the dependencies between them, 
        and their usage, are taken into account.
    % Moreover,  
    %     as we don't have to deal with multiple models,
    %     this approach is potentially cheaper in computation time.
    Our architecture is comprised of a CDE model, 
        which joins a CVAE \cite{Kingma2013AutoEncodingVB, Sohn2015LearningSO} 
        and a CNF model \cite{Dinh2014NICENI, Kingma2018GlowGF, Dinh2016DensityEU}.

%% file: content/background/background.tex
\section{Background} \label{sec:background}

    % This section discusses recent efforts in Conditional Variational Autoencoders (CVAEs) and Conditional Invertible Normalizing Flows(CNF).
    
    \input{content/background/cvae}
    \input{content/background/cnf}

%% file: content/background/cvae.tex
\subsection{Conditional Variational Autoencoders} \label{sec:cvae}

    VAE \cite{Kingma2013AutoEncodingVB} is a probabilistic model that estimates the marginal distribution $p(x)$ of the observations $x \in X$ through a lower bound constructed with a posterior distribution approximation \cite{Jordan1999AnIT}. 
    CVAE is a  VAE extension to the conditional case.
    Thus,
        CVAE extends the VAE's math by conditioning the entire generative process on an input $y$ \cite{Sohn2015LearningSO, Walker2016AnUF, Doersch2016TutorialOV}. Given an input $y$ and an output $x$, 
        a CVAE aims at creating a model $p( x | y )$, 
        which maximizes the probability of ground truth.
        % which maximizes the ground truth's probability. 
    The model is defined by introducing a latent variable 
        $z \sim \mathbb{N}( 0, I )$ 
        and a lower bound on the log marginal probability. 
        \begin{equation}
            \label{eq:elbo-cvae}
            log\ p_{\theta}(x|y) \geq \mathbb{E}_{q} \left [\ log\ p_{\theta}(x|z,y)\ \right ] - \mathbb{KL}(q_{\phi}(z|x,y) || \pi(z|y)) 
        \end{equation}
    Equation \ref{eq:elbo-cvae} is refered to as 
        the Evidence Lower Bound (ELBO) to the conditional case. 
    In (1), 
        $q_{\phi}(z|x,y)$ is the posterior approximation,
        $p_{\theta}(x|z,y)$ is a likelihood function,
        $\pi(z|y)$ is the prior distribution,
        and $\phi$ and $\theta$ are the model parameters.
        
    %%%%%%%%%%%%%%%%%%%%%%%%%%%%%%%%%%%%%%%%%%
    % VAE and CVAE models usually results in biased maximum likelihood estimates \cite{Berg2018SylvesterNF} due to their formulation choices for the posterior approximation.
    % At last, 
    %     it is worth mentioning that the better the posterior approximation, 
    %     the tighter the ELBO is, 
    %     and so smaller is the gap between the true distribution $p(x)$ and its lower-bound.

%% file: content/background/cnf.tex
\subsection{Conditional Normalizing Flow Models} \label{sec:conditional-flow}
    NF models 
    \cite{Dinh2014NICENI, Dinh2016DensityEU, Kingma2018GlowGF} 
        learn flexible distributions by transforming a simple base distribution with invertible transformations, 
        known as normalizing flows. 
    CNF \cite{Ardizzone2019GuidedIG, Winkler2019LearningLW} is an NF extension to the conditional case. 
    Given an input $y$ and a target $x$,
        we learn the distribution $p(x|y)$ 
        using the conditional base distribution $\pi(z_{k}|y)$ 
        and a mapping $z_{k} = f_{\omega}(x,y)$
        which is bijective in $x$ 
        and $z_{k}$.
    In such model,
        the inputs probability is maximized through the change of variables formula:
            \begin{equation}
                \label{eq:conditional-change-of-variable-formula}
                p(x|y) = \pi(f_{\omega}(x,y)|y)\left | det \left (\frac{\partial f_{\omega}(x,y)}{\partial x} \right ) \right |
            \end{equation}
    
    In equation \ref{eq:conditional-change-of-variable-formula}, 
        the term $| det (\frac{\partial f_{\omega}(x,y)}{\partial x}  ) |$
        is the Jacobian Determinant absolute value of $f_{\omega}$.
        % or for simplicity, the Jacobian of $f_{\omega}$.
    It measures the change in density when going from $x$ to $z$ under the transformation $f_{\omega}$.
    
    The main challenge in invertible normalizing flows is to design mappings to compose the transformation $f_{\omega}$ since they must have a set of restricting properties \cite{Winkler2019LearningLW}. 
    % Thus,
    %     all mappings $f^{i} \in f_{\omega}$ must have 
    %     a known and tractable inverse mapping  
    %     and an efficiently computable Jacobian,
    %     while being powerful enough to model complex transformations.
    % Moreover, 
    %     fast sampling is desired, 
    %     so the inverse mappings $(f^{i})^{-1}$ should be calculated efficiently. 
    Fortunately, 
        there is a set of well-established invertible layers with all those properties on the current normalizing flows literature.
    This work is built upon three of them
    \footnote{Due to space constraints we refer to \cite{Kingma2018GlowGF} for details on invertible layers.}: Affine Coupling Layer (ACL)
            \cite{Dinh2014NICENI,Dinh2016DensityEU};  
            Invertible $1\times1$ Convolution layer \cite{Kingma2018GlowGF}; 
           Actnorm layer \cite{Kingma2018GlowGF}.
    
    At last,
        the conditional data $y$ is encoded by a network into a rich representation $h(y)$,
        and then is introduced  
        (I) in the ACL
        and (II) in the base distribution. 
    For the ACL,
        we pass the conditional information $h(y)$ by concatenating it with the inputs of the layer's internal operators,
        such as proposed by Ardizzone \cite{Ardizzone2019GuidedIG}. 
    For the base distribution,
        the conditional data is introduced in its parameters calculation, such that:
            $\pi(z_{k}|y) = \mathcal{N}(z; \mu(h(y)), \sigma(h(y)))$,
        where $\mu(h(y))$ and $\sigma(h(y))$ are the outputs of a standard neural network that uses the conditioning term $h(y)$ as input.

%% file: content/cfvae/cfvae.tex
\section{\ModelFullName} \label{sec:pfvae}
    
    This section discusses our proposed {\ModelName} model and motivations.
    In the following we denote the aggregate data as $y$ 
        and the individual appliances data as $x$.  
    Our approach is to model NILM to estimate a noisy distribution 
        which is conditioned on the aggregated power demand.
    For this purpose, 
        one could simply model the conditional distribution $p(x|y)$ directly using a CVAE model.
    However, 
        CVAEs assume a diagonal-covariance Gaussian on the latent variables, 
        which induces a strong model bias 
        making it a challenge to capture multi-modal distributions \cite{Berg2018SylvesterNF, Bhattacharyya2019ConditionalFV}.
    This also leads to missing modes due to posterior collapse \cite{Ziegler2019LatentNF}.
    On the other hand,
        CNFs allow tracktable exact likelihood estimation, 
        in contrast to VAEs' lower bound estimation.
    However, 
        CNFs also have its limitations.
    The bijective nature of the transformations used 
        for building NFs limits their ability to alter dimensionality, 
        and to model structured data, 
        and distributions with disconnected components \cite{Nielsen2020SurVAEFS}.
    CVAEs have no such limitations.
    Several methods for mitigating such limitations of CVAEs and CNF have been proposed \cite{Berg2018SylvesterNF, Bhattacharyya2019ConditionalFV, Ziegler2019LatentNF, Yang2019PointFlow3P, Ma2019FlowSeqNC, Nielsen2020SurVAEFS} and our proposal is built upon their ideas.

    We propose a CDE model,
        which joins CVAEs 
        with CNFs
        to estimate the conditional power demand density for each appliance.
      The resulting {\ModelName} model is a CVAE extension 
        that uses a CNF model to learn the prior distribution $\pi(z_{0}|y)$. 
    Therefore, 
        the {\ModelName} is composed of two components. 
    The first component is the CVAE that still learns 
        an inference $q_{\phi}(z_{0} | x,y)$ 
        and a generative $p_{\theta}(x | z_{0}, y)$ model. 
    The second component is a CNF model responsible for learning the prior distribution $\pi(z_{0} | y)$.
   
    During the training procedure,
        the inference model outputs $z_{0}$.
    After that, 
        $z_{0}$ is passed throughout the CNF model, 
        which encodes $z_{0}$ into $z_{k}$.  
    Concurrently,
        $z_{0}$ is also given as input to the decoder model,
        decoding $z_{0}$ back to $x$. 
    The whole model is jointly trained using stochastic gradient descent with mini-batches.
    In this setting,
        our ELBO manipulation becomes:
        % \input{content/cfvae/equations/eq_cfvae_elbo}
        \input{content/cfvae/equations/eq_cfvae_lower_bound}%
        where $\omega$, 
        $\phi$ 
        and $\theta$ are the model's trainable parameters.
    % Thus,
    %     by passing the log-posterior expectation to the left side of equation \ref{eq:elbo-cfvae},
    %     we define a new lower-bound for the model optimization: 
    %     \input{content/cfvae/equations/eq_cfvae_lower_bound}%
    The loss function has two terms: 
        the reconstruction error; 
        and the regularization term.
    % Since the decoder targets $x$ are in the continuous space,
    %     we consider that the decoder is 
    %     a multivariate Gaussian with a diagonal covariance structure.
    We use the commonly chosen mean squared error simplification for the reconstruction error term.
    
    The generative process is done by 
        first sampling $z_{k}$ from a simple base density,
        whose parameters $\mu(h(y))$ and $\sigma(h(y))$ are calculated by the conditioning network. %using the $y$ observation.
    The conditioning network also outputs a hidden state $h(y)$,
        which is passed as input to the inverse mapping alongside with $z_k$ to recover the latent variable $z_{0}$. %such that $z_{0} = f^{-1}(z_{k}, h(y))$. 
    % Thus,
    %     the conditioning network has three outputs
    %     $\mu(h(y))$, 
    %     $\sigma(h(y))$,
    %     and $h(y)$. 
    At last,
        $z_{0}$ is decoded by the generative model $q_{\phi}(x | z_{0}, h(y))$.

    Figure \ref{fig:cfvae-train-and-inference-schemes}
        illustrates the model's training and inference schemes.
    In (a), we present the training flow scheme, 
        and we present the inference flow scheme in (b).
    % In (a), 
        % we present the training flow scheme. %, 
        % where the outputs $\hat{x}$ 
        % and $z_{k}$ are used to compute the loss function, 
        % such as defined by equation \ref{eq:cfvae-lower-bound}.
    % In (b), 
        % we present the inference flow scheme., 
        % where we use the conditioning network outputs,
        % $\mu(h(y))$ and  $\sigma(h(y))$, 
        % to sample $z_{k}$ through the reparametrization trick 
        % and passing it through the CNF inverse pass to recover $z_{0}$.
    Note that,
        in both (a) and (b),
        we concatenate $z_0$ and $h(y)$ 
        into a single tensor before serving them as inputs to the decoder network, 
        but the inputs are not concatenated in the CNF.
    It is omitted from the diagrams for simplicity.
    \input{content/cfvae/figures/fig_cfvae_schemes}%

%% file: content/cfvae/equations/eq_cfvae_lower_bound.tex
% \begin{equation}
\begin{multline}
    \label{eq:cfvae-lower-bound}
    log\ p_{\theta}(x|y) 
    + \mathbb{E} \left [\ log\ q_{\phi}(z_{0}|x,y)\ \right ] \\
    \geq 
        \mathbb{E} \left [\ log\ p_{\theta}(x|z_{0},y)\ \right ]
        + \mathbb{E} \left [\ log\ \pi_{\omega}(z_{0}|y)\ \right ]
\end{multline}%
% \end{equation}
%

%% file: content/cfvae/figures/fig_cfvae_schemes.tex
    \begin{figure}[h]
        \centering
        \begin{minipage}{7cm}
            \includegraphics[height=6cm, width=6cm]
            {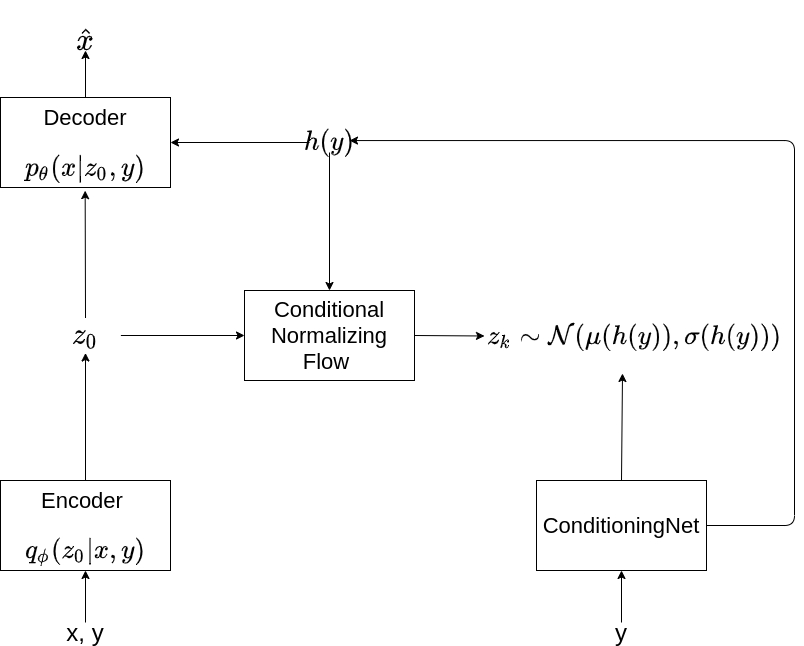}
            \caption*{
                (a) {\ModelName} \space training scheme.
            }
            \label{fig:cfvae-train-scheme}
        \end{minipage}
        \qquad
        \begin{minipage}{7cm}
            \includegraphics[height=6cm, width=6cm]
            {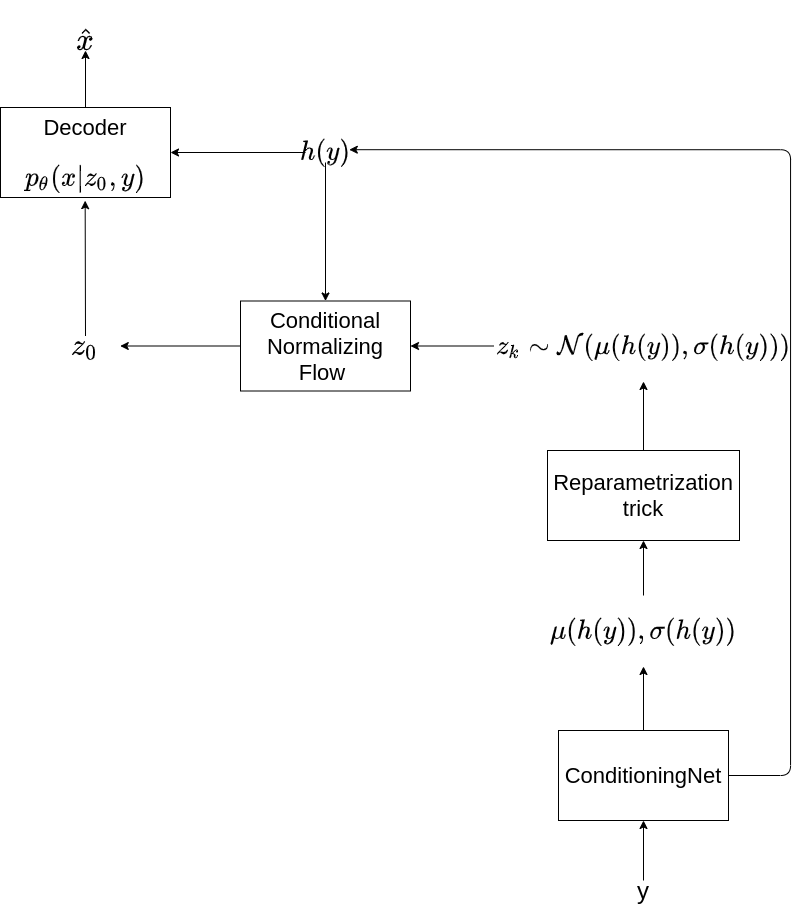}
            \caption*{
                (b) {\ModelName} \space inference scheme.
            }
            \label{fig:cfvae-test-scheme}
        \end{minipage}
        \caption{
                {\ModelName}'s train and inference schemes.
            }
            \label{fig:cfvae-train-and-inference-schemes}
    \end{figure}%

%% file: content/architecture-details/architecture-details.tex
\section{Architecture Details} \label{sec:architecture}

        \input{content/architecture-details/encoder_and_conditioning_networks}
        
        \input{content/architecture-details/decoder_network}
        
        \input{content/architecture-details/CNF_Network}

%% file: content/architecture-details/encoder_and_conditioning_networks.tex
\begin{figure}[h]
            \centering
            \includegraphics[width=\linewidth]
            {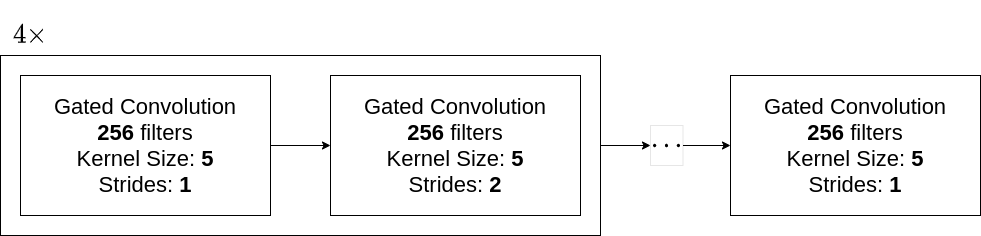}
            % \begin{tikzcd}
            %     \text{Gated Convolution} \arrow{r} &%\\%& 
            %     \text{Gated Convolution}
            % \end{tikzcd}
            \caption{Encoder and Conditioning networks.}
            \label{fig:gated-block}
        \end{figure}

    Both the Encoder and Conditioning networks are simple gated convolutional neural networks \cite{Dauphin2017LanguageMW}. 
    They are both composed by {\NEncoderBlocks} gated blocks, 
        illustrated in Figure \ref{fig:gated-block}. 
    At each of these blocks, 
        the temporal dimension is halved by the second gated convolution block.
    
    Additionally, 
        these two networks only differ in the final layer. 
    After the last Gated Convolution, 
        two convolutional layers in the Conditioning network
        receive the hidden state $h(y)$, 
        outputting $\mu(h(y))$ and $\sigma(h(y))$.
    These convolutional layers have 
        {\Zsize} filters 
        of size {\EncoderLastKSize} 
        and stride $1$.
    The encoder network has one additional convolutional layer 
        % after the last Gated Convolution with {\Zsize} filters 
        with {\Zsize} filters 
        of size {\EncoderLastKSize} 
        and stride $1$.
        
    Therefore,
        % the Encoder receives the concatenated vectors $[x;y]$ as inputs
        the Encoder receives the concatenated vectors $[x;y]$
        % and encodes them into a latent representation $z_{0}$, 
        and outputs the latent representation $z_{0}$, 
        % whereas the Conditioning network receives $y$ as input,
        whereas the Conditioning network receives $y$,
        encodes it into a rich representation $h(y)$
        and into the base distribution parameters $\mu(h(y))$ and $\sigma(h(y))$. 

%% file: content/architecture-details/decoder_network.tex
%%\subsubsection*{Decoder Network}

    The Decoder network is a straightforward neural network 
        % responsible for decoding the examples from the latent space $z_{0}$ back to the observation space $x$.
        responsible for decoding $z_{0}$ back to the $x$ space.
     
    Thus, 
        the Decoder Network is composed of {\NDecoderBlocks} decoder blocks followed by two simple convolutional layers.
    \begin{figure}[h]
        \centering
        \includegraphics[width=\linewidth]
            {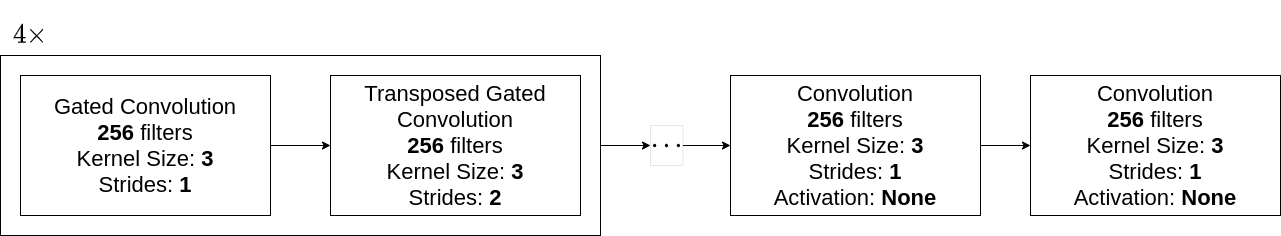}
        \caption{Decoder network.}
        \label{fig:decoder-block}
    \end{figure}

%% file: content/architecture-details/CNF_Network.tex
    The CNF network is responsible for learning the prior distribution $\pi(z_{0}|y)$. 
    This network consists of {\NCNFBlocks} stacked step-flow blocks \cite{Kingma2018GlowGF}, 
        which, in it turns, 
        is a stacking of 
        an Actnorm layer, 
        an Invertible $1 \times 1$ Convolution layer, 
        and an ACL.
    
\begin{figure}[h]
            \centering
            \includegraphics[width=\linewidth]
            {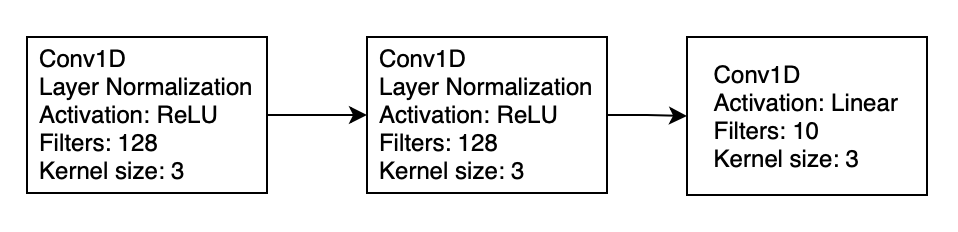}
            \caption{Coupling Layer's backbone}
            \label{fig:s-and-t-block}
        \end{figure}

   The ACL operators are calculated by a backbone network, 
        which is illustrated in Figure \ref{fig:s-and-t-block}. 

%% file: content/experiments/experiments.tex
\section{Experiments} \label{sec:experiments}

 Our experiments were conducted on a publicly available dataset \cite{Martins2018Dataset} 
         that contains industrial heavy-machinery data collected at a Brazilian poultry feed factory.     
    The dataset provides measurements individually collected from eight machines 
        and the factory's main voltage distribution board (the aggregated data). 
    The available data consists of time series, 
        sampled at 1 Hz, 
        of the following electrical quantities:
        the RMS voltage, 
        the RMS current, 
        the active power, 
        the reactive power, 
        the apparent power, 
        and the active energy.
    
    The appliance measurements are processed to create an overlapping grid 
        of intervals used as inputs and ground-truth targets by our model.
    The windows width was decided during our preliminary experiments. %with the dataset. 
    We found that intervals of size {\GirdSize} are long enough 
        to capture most machinery activations. %for all dataset’s machinery. 
    Increasing or decreasing the window size 
        could improve some individual machinery results. 
    Still, 
        since our goal is to work with a single model, %for all machinery, 
        intervals of size {\GirdSize} presented promising results.
    
    The resulting dataset comprises {\DatasetSize} appliance samples. 
        % and their respective aggregated electrical quantities.
    Figure \ref{fig:power-demand-appliances} presents an aggregated appliance interval 
        and its respective per-machinery active power  demand (the targets) taken from the resulting dataset. 
    The data is normalized per-channel by the mean and standard deviation.
    \input{content/dataset/figures/fig_power_appliances}

    In all experiments, 
        the model performance is evaluated using two commonly used metrics in NILM tasks \cite{zhong2014signal}: 
        the normalized disaggregation error (NDE) 
        and the signal aggregated error (SAE). 
    % The NDE is used to verify a NILM model's capability in predicting the appliance's instantaneous power demand.
    % In its turn, 
    %     the SAE is used to verify the model's ability in predicting the appliance's total energy consumption.
    Those metrics are calculated from the average of $20$ samples taken with the trained models using the aggregated data $y$ as input.
    
    All models are trained via gradient descent with mini-batches of size $50$ to maximize the equation \ref{eq:cfvae-lower-bound}. 
    %minimize the loss described by equation \ref{eq:cfvae-lower-bound}.
    Thus, 
        the {\ModelName} is trained on a Google Cloud TPU v2-8 node for $2,000$ epochs
        using Adam optimizer \cite{Kingma2015AdamAM} with its default parameters
        and a learning rate of {\LearningRate}.
    
    \input{content/experiments/cross_validation}
    
    \input{content/experiments/ablation}

%% file: content/dataset/figures/fig_power_appliances.tex
    \begin{figure}[h]
        \centering
        \begin{minipage}{7cm}
        % \begin{minipage}
        % \parbox{5cm}{
            \includegraphics[height=6cm, width=7cm]
            {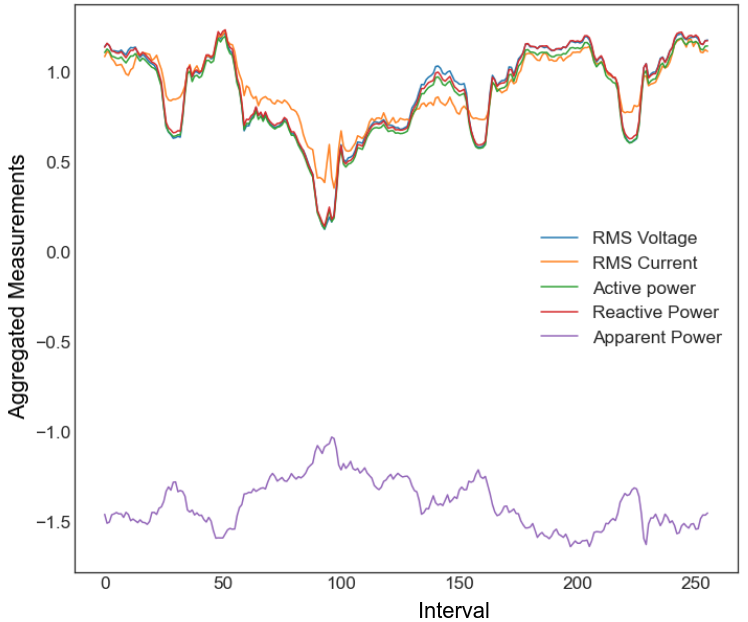}
            \caption*{
                (a) Aggregate measurements' features.  
            }
            % \label{fig:cfvae-train-scheme}
        % }
        \end{minipage}
        \qquad
        \vskip 0.15in
        \begin{minipage}{7cm}
        % \begin{minipage}
            \includegraphics[height=6cm, width=7cm]
            {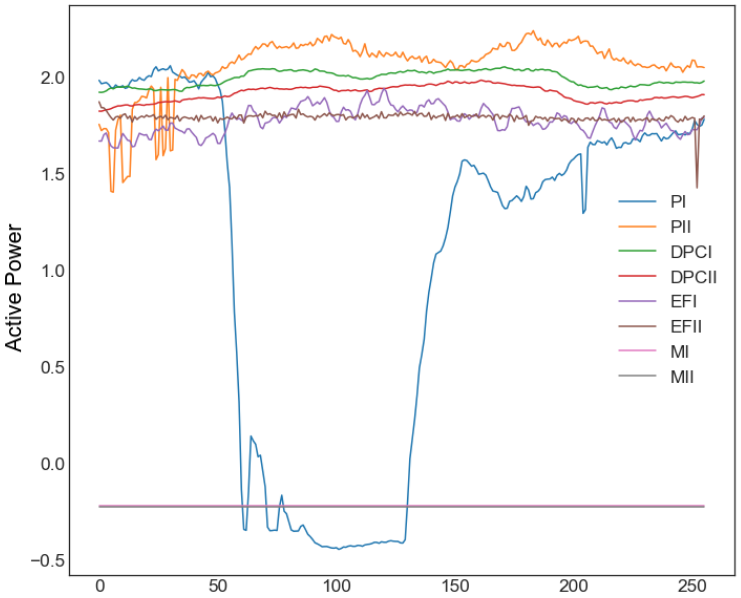}
            \caption*{
                (b) Appliance-by-appliance power demand.  
            }
            % \label{fig:cfvae-test-scheme}
        \end{minipage}
        \caption{
          Example of a window of data present in the dataset.  
        }
        \label{fig:power-demand-appliances}
    \end{figure}

%% file: content/experiments/cross_validation.tex
\subsection*{Cross-Validation}  \label{sec:results} 
    
    This experiment is performed in a {\NFolds}-fold cross-validation (CV) setting \cite{CV:Kohavi:1995}.
    The training procedure takes on average 17 hours per cross-validation iteration. 
    Additionally,
        we perform the CV apart for the milling machines (MI and MII) because the available data is smaller for them.
        % since the amount of available data for the milling machines (MI and MII) is much smaller than the available data for the other appliances,
        % we perform the CV apart for these machines.
    Table \ref{tab:results} compares the obtained metrics with the WaveNet (WN) model , 
        proposed by the previous work \cite{Martins2018ApplicationOA} on the same dataset.
        
    \input{content/experiments/tables/results_table}
    Table 1 shows that the {\ModelName} consistently improves the metrics in six of the eight machines belonging to the dataset.
    Analyzing the NDE metric, 
        the improvements go from {\PINDEImprovement}, 
        for the \textit{PI} machinery, 
        up to {\MINDEImprovement}, 
        for the \textit{MI} machinery.
    Additionally, 
        there is an error increase for the \textit{EFI} and \textit{EFII} machinery. 
    In turn, 
        improvements in SAE go from {\PIISAEImprovement} 
        for the \textit{PII} machinery 
        up to {\DPCIISAEImprovement} for the \textit{DPCII} machinery.
    We also observe an error increase for the \textit{EFI},
        \textit{EFII}, and \textit{MII} machinery.
    
    % Finally, 
    %     the PFVAE's higher representational capacity enables the usage of all electrical quantities available in the dataset as input 
    %     % features 
    %     that might explain the model's overall improvements.
    The appliance activations are generally related to each other due to the factory's operation routine,
    and therefore, 
    learning to represent multiple appliance activations using a single model can be beneficial.
    Our experiments reinforce this hypothesis, 
        which together with the PFVAE's higher representational capacity might explain the model's overall improvements.
    
    % Moreover, 
    %     due to the factory's operation routine, 
    %     the appliance activations are generally related to each other. 
    % Therefore learning to represent multiple appliance activations using a single model can be beneficial.

%% file: content/experiments/tables/results_table.tex
% \newcolumntype{R}{>{\raggedleft\arraybackslash}X}%
% \newcolumntype{L}[1]{>{\raggedright\arraybackslash}X}%
\begin{table}[ht]
    \caption{
        Performance comparison between the {\ModelName} and the Wavenet (WN) models.
    }
    \label{tab:results}
    \begin{center}
    \begin{small}
    \begin{sc}
    \begin{tabularx}{0.8\columnwidth}{X ll}
        &
            \multicolumn{2}{c}{NDE} \\
        &  
             \multicolumn{1}{c}{WN} 
            %  WN 
        & 
            \multicolumn{1}{c}{{\ModelName}} \\   
            % {\ModelName} \\   
        \bottomrule
        \midrule
        
        PI    & 
            0.045 $\pm$ 0.002  &  {\ndePI} \\
            
        PII    & 
            0.056 $\pm$ 0.002 & {\ndePII} \\
                
        DPCI    & 
            0.08 $\pm$ 0.01  & {\ndeDPCI} \\
        
        DPCII    & 
            0.202 $\pm$ 0.006  & {\ndeDPCII} \\
        
        EFI    & 
            \textbf{0.046 $\pm$ 0.0004}  & {\ndeEFI} \\
            
        EFII    & 
            \textbf{0.041 $\pm$ 0.003}  & {\ndeEFII} \\
            
        MI    & 
            0.08 $\pm$ 0.02  & {\ndeMI} \\
        
        MII    & 
            0.06 $\pm$ 0.01  & {\ndeMII} \\
        \\    
        % \midrule 
        % \toprule
        &
            \multicolumn{2}{c}{SAE} \\
        &  
             \multicolumn{1}{c}{WN}
        & 
            \multicolumn{1}{c}{{\ModelName}} \\
        \bottomrule   
        \midrule
        
        PI    & 
            0.047 $\pm$ 0.009 & {\saePI} \\

        PII    & 
            0.022 $\pm$ 0.009 & {\saePII} \\
                
        DPCI    & 
            0.13 $\pm$ 0.04  & {\saeDPCI} \\
        
        DPCII    & 
            0.19 $\pm$ 0.02  & {\saeDPCII} \\
        
        EFI    & 
            \textbf{0.007 $\pm$ 0.006}  & {\saeEFI} \\
            
        EFII    & 
            \textbf{0.016 $\pm$ 0.009}  & {\saeEFII} \\ 
            
        MI    & 
            0.09 $\pm$ 0.04 & {\saeMI} \\
            
        MII    & 
            \textbf{0.03 $\pm$ 0.02} & {\saeMII} %\\
            
    \end{tabularx}
    \end{sc}
    \end{small}
    \end{center}
    % \vskip -0.15in
\end{table}

%% file: content/experiments/ablation.tex
\subsection*{Ablation Studies}  \label{sec:ablation}
    In the {\ModelName} architecture, 
        we included two components that work for conditioning the generative process on the aggregate data $y$, 
        which are 
        the $h(y)$ connections in the ACL and 
        % the conditioning from $z_{k}$ base distribution in the CNF.
        the conditioning from the CNF base distribution.
    We conducted two experiments to investigate the contribution of these components.
    % Therefore, 
    %     to investigate the contribution of these components', % to the model, 
    %     we conducted two experiments consisting of training and testing {\ModelName}
    %     variations without each of these components. 
    
    In both experiments,
        the data and the model hyper-parameters were kept constant. 
        % while different {\ModelName} versions were evaluated.   
    The models were trained for six of the eight machines, 
        using $80\%$ of the dataset. 
    The remaining $20\%$ of the dataset was used for testing.
    % Additionally,
    We present the total sum and the averaged metrics over the six machines
        % \footnote{We present the individual results in the supplementary material.
        % https://arxiv.org/abs/2011.14870}.
        \footnote{We present the individual results in the appendices}.
    % Thus,
    %     the metrics are computed individually for each machine, 
    %     and then the total sum and the average values are calculated.
    
    First,
        we compared the complete {\ModelName} model's performance against two modified versions.
    % In the first modified version,
    In the first version, 
        we removed the $h(y)$ connections 
        from the ACLs 
        by exchanging them for its simple (and original) version.
    % In the second modified version, 
    In the second version,
        we removed the 
            $\mu(h(y))$ and $\sigma(h(y))$ 
            calculations from the model,
            fixing the CNF base distribution such that $\pi(z_{k}|y) \sim \mathcal{N}(0,1)$.
        % we removed the conditioning from the CNF base distribution, 
        % by fixing its parameters to match the diagonal standard normal distribution such that $\pi(z_{k}|y) \sim \mathcal{N}(0,1)$. 
    % As a consequence,
    %         we removed the 
    %         $\mu(h(y))$ and $\sigma(h(y))$ 
    %         calculations from the model.
        
    \input{content/experiments/tables/conditioning_ablations_table}
    
    Table \ref{tab:conditioning-ablations} compares the averaged 
        and total resulting metrics of the 
        complete model, 
        the simple ACL,
        and the standard normal base distribution versions. 
    The standard normal base distribution version results 
        were around ten times worse in both metrics than the other versions.%,
        % which might indicate that 
        % learning the base distribution parameters allows 
        % the CNF to estimate tighter and more flexible distributions for $z_{0}$.
    In it turns,
        the simple ACL version presents very close values for the SAE metric compared against the complete model.
    % Comparing the simple ACL version 
    %     against the complete model indicates that 
    %     both versions' capability in predicting the overall energy consumption 
    %     for the appliances is quite similar 
    %     since the SAE for these two versions is very close.
    However, 
        the significant difference in the NDE indicates that
        the complete model has a stronger capability
        to predict appliance instantaneous power demand.
    Therefore, 
        the usage of the $h(y)$ connections 
        in the ACLs seem to be 
        necessary for the model's performance.%, 
        % which justifies its use, 
        % even though it increases the total number of parameters.
    
    Next, %we performed another ablation study, 
       we analyse  the {\ModelName} model's performance 
    %   analyzing the {\ModelName} model's performance 
        when varying the number of step-flow blocks used in the CNF.
    Table \ref{tab:cnf-ablations} compares 
        the averaged and total resulting metrics 
        for five trained model variants with 
        $2$, $4$, $8$, $16$, and $32$ 
        step-flow blocks.% in the CNF component.
        
    \input{content/experiments/tables/cnf_ablations_table}
    
    The model with $8$ step-flow blocks presented the best performance in both metrics. 
    It is worth noting that the performance improves by adding more blocks until $8$, 
        starting to degrade when adding more blocks.%, 
        % as seen in the results for $16$ and $32$ blocks.
    Thus,
        the model's performance degrades with a too shallow and a very deep CNF. 
    % It indicates that we must choose the number of step-flows sparingly 
        % and that the model's performance degrades with very deep CNFs, 
        % even though a too shallow model is not powerful enough to capture the latent space distribution. 
        
    In summary, 
        these experiments suggest that the best version of the {\ModelName} for the task 
        % is the one where the CNF is built with $8$ step-flow blocks, 
        is the complete model with $8$ step-flow blocks. 
        % uses learned features $h(y)$ in its ACLs,
        % and is conditioned on the $z_k$ base distribution.

%% file: content/experiments/tables/conditioning_ablations_table.tex
% \newcolumntype{R}{>{\raggedleft\arraybackslash}X}%
% \newcolumntype{L}{>{\raggedright\arraybackslash}X}%
\begin{table}[ht]
    \caption{
        Comparison of the averaged and total performance between ablation models.
        % Aggregated and Averaged performance comparison between {\ModelName} ablations.
    }
    \label{tab:conditioning-ablations}
    \begin{center}
    \begin{small}
    \begin{sc}
    \begin{tabularx}{0.8\columnwidth}{L cc}
        % &
        % \toprule
        \multicolumn{3}{c}{Total} \\
        Ablation
        & 
        SAE 
        &  
        NDE \\
        \bottomrule
        \midrule
        
        \textbf{Complete Model}
            & 
            \textbf{0.185}  &  \textbf{0.373} \\%&
            
        Simple Affine Layer
            & 
            0.200 & 0.452 \\%&
        
        % Constant $z_k$ prior distribution.
        Standard Normal
            & 
            2.470  & 5.301 \\%&
        
        % \midrule 
        
        % &
        % \toprule
        \\
        \multicolumn{3}{c}{Averaged} \\
        Ablation
        % \multirow{2}{4em}{Metric/Models} &
        & 
        SAE
        &  
        NDE  \\
        \bottomrule
        \midrule
        
        \textbf{Complete Model}
            & 
            \textbf{0.030}  &  \textbf{0.062} \\%&
            
        Simple Affine Layer
            & 
            0.033 & 0.075 \\%&
        
        % Constant $z_k$ prior distribution.
        Standard Normal
            & 
            0.411  & 0.883 \\%&
        % \bottomrule
    \end{tabularx}
    \end{sc}
    \end{small}
    \end{center}
    \vskip -0.15in
\end{table}

%% file: content/experiments/tables/cnf_ablations_table.tex
% \newcolumntype{R}{>{\raggedleft\arraybackslash}X}%
% \newcolumntype{L}[1]{>{\raggedright\arraybackslash}X}%
\begin{table}[ht]
    \caption{
        Comparison of the averaged and total performance between {\ModelName} variants with 2, 4, 8, 16, and 32 step-flow blocks in the CNF component.
    }
    \label{tab:cnf-ablations}
    \begin{center}
    \begin{small}
    \begin{sc}
    \begin{tabularx}{0.8\columnwidth}{L cc}
        % \toprule
        % &
        \multicolumn{3}{c}{Total} \\
        Step-Flows
        & 
        SAE 
        &  
        NDE \\
        \bottomrule
        \midrule
        
        2
            & 
            1.189  &  2.317 \\%&
            
        4
            & 
            0.261 & 0.611 \\%&
        
        \textbf{8}
            & 
            \textbf{0.185}  & \textbf{0.373} \\%&
        
        16
            & 
            0.431  & 0.663 \\%&
        
        32
            & 
            1.598  & 4.196 %\\%&
        
        % \midrule 
        % \toprule
        \\
        \multicolumn{3}{c}{Averaged} \\
         
        Step-Flows
        & 
        SAE 
        &  
        NDE \\
        
        \bottomrule
        \midrule
        
        2
            & 
            0.198  &  0.386 \\%&
            
        4
            & 
            0.043 & 0.101 \\%&
        
        \textbf{8}
            & 
            \textbf{0.030}  & \textbf{0.062} \\%&
        
        16
            & 
            0.071  & 0.110 \\%&
        
        32
            & 
            0.266  & 0.699 %\\%&
    \end{tabularx}
    \end{sc}
    \end{small}
    \end{center}
    \vskip -0.1in
\end{table}

%% file: content/conclusion.tex
\section{Conclusions} \label{sec:conclusion}
    This work proposes a CDE model 
        to estimate the individual appliances' power demand 
        % from the aggregate electrical quantity measurements observations.
        from their aggregate power demand measurements.
    The proposed model joins a CVAE 
        with a CNF model.%, 
        % which estimates the conditional prior distribution 
        % based on the posterior and prior data observations.
        
    We conducted experiments using a publicly available dataset
        \cite{Martins2018Dataset} 
        with data collected from a Brazilian poultry feed factory.
    We observed improvements ranging from 
        {\PINDEImprovement} up to {\MINDEImprovement} in NDE
        and {\PIISAEImprovement} up to {\DPCIISAEImprovement} in SAE
        for six dataset machines.
    
    Future work include testing our model using datasets with different appliances 
        and scenarios, 
        such as commercial and residential buildings datasets like 
        \cite{UK-DALE}\cite{kolter2011redd}.
        % researching ways to accelerate the model convergence,
        % and refinements aimed for higher accuracy.
    Additionally, 
        natural gas and water usage may also be monitored by similar methods \cite{Hart1992NonintrusiveAL}.
    
    % Finally,
    %     our experimental results show that the proposed model achieves performance comparable to the previous best results in the same dataset %in the NILM task, 
    %     improving the disaggregation performance for most appliances belonging to the dataset. 
    Finally,
        our experimental results show that the proposed model improves the disaggregation performance for most appliances belonging to the dataset.

%% file: supplementary_material/main_supplementary.tex
\appendices
\onecolumn
% \title{ Supplementary material for the paper \\
%                 {\ModelFullName}: A density estimation model for Non-Intrusive Load Monitoring
% }

% \author{
%     Luis Felipe M.O. Henriques \\
%       Department of Computer Science\\
%       Pontifical Catholic University of Rio de Janeiro\\
%       Rio de Janeiro, Brazil \\
%       \texttt{lhenriques@inf.puc-rio.br} \\
%   \AND
%   Eduardo Morgan \\
% %   \thanks{Use footnote for providing further
%     % information about author (webpage, alternative
%     % address)---\emph{not} for acknowledging funding agencies.} \\
%       Department of Computer Science\\
%       Pontifical Catholic University of Rio de Janeiro\\
%       Rio de Janeiro, Brazil \\
%       \texttt{emorgan@inf.puc-rio.br}
%   \And
%   Sergio Colcher \\
%       Department of Computer Science\\
%       Pontifical Catholic University of Rio de Janeiro\\
%       Rio de Janeiro, Brazil \\
%       \texttt{colcher@inf.puc-rio.br}
%       %% examples of more authors
%   \And
%   Ruy Luiz Milidiú \\
% %   \thanks{Use footnote for providing further
%     % information about author (webpage, alternative
%     % address)---\emph{not} for acknowledging funding agencies.} \\
%   Department of Computer Science\\
%   Pontifical Catholic University of Rio de Janeiro\\
%   Rio de Janeiro, Brazil \\
%   \texttt{milidiu@inf.puc-rio.br} \\
% }

% \begin{document}
% \maketitle
% \input{supplementary_material/abstract}
    
\input{supplementary_material/experiments/experiments}

%% file: supplementary_material/experiments/experiments.tex
\section{Ablation Studies Details} \label{apdx:ablations}
    
    \input{supplementary_material/experiments/ablation}
    \input{supplementary_material/experiments/step-flows-ablation}

\section{Visual Comparison Plots for Qualitative Analysis} \label{apdx:comparison-plots}    
    \input{supplementary_material/experiments/Visual-Comparison-Plots}

%% file: supplementary_material/experiments/ablation.tex
\subsection*{Conditioning Ablations}  \label{apdx:conditioning-ablations}
    Section \ref{sec:experiments} of the paper presents ablation studies in the {\ModelName}'s components responsible for conditioning the generative process in the aggregate data $y$. 
    Three models were trained in the experiment:
    \begin{enumerate}
        \item \textbf{Simple Affine Layer} - 
            We removed the conditioning $h(y)$ connections from the affine coupling layers by exchanging them for their original version. 
            Figure \ref{fig:simple-affine-train-and-inference-schemes} illustrates the Simple Affine Layer ablation training  and  inference schemes.
            \input{supplementary_material/experiments/figures/fig-simple-affine-ablation-schemes}
            
        \item \textbf{Standard Normal} - 
            We removed the conditioning from the $z_{k}$ base distribution,
            in the CNF, 
            by fixing its parameters to match the diagonal Standard Normal such that $\pi(z_{k}|y) \sim \mathcal{N}(0,I)$.
            Consequently, we removed the 
            $\mu(h(y))$ and $\sigma(h(y))$ calculations from the model.
            Figure \ref{fig:standard-normal-train-and-inference-schemes} illustrates the Standard Normal ablation training  and  inference schemes.
            \input{supplementary_material/experiments/figures/fig-standard-normal-ablation-scheme}
            
        \item \textbf{Full {\ModelName} Model} -
            This variant is the complete {\ModelName} model with no modifications.
            Figure \ref{fig:full-model-train-and-inference-schemes} illustrates the Full {\ModelName} Model ablation training  and  inference schemes. 
            Note, 
                this figure is the same one presented in the paper's section \ref{sec:pfvae}, 
                and we are showing it here again for easier comparison.
            \input{supplementary_material/experiments/figures/fig-full-ablation-scheme}
    \end{enumerate}
    
    Table \ref{tab:cnf-ablations-per-machine} compares the models' performance per machine.
    \input{supplementary_material/experiments/tables/conditioning_ablations_table}
    
\begin{figure}[h]
        \centering
        \includegraphics[width=0.9\textwidth]
        {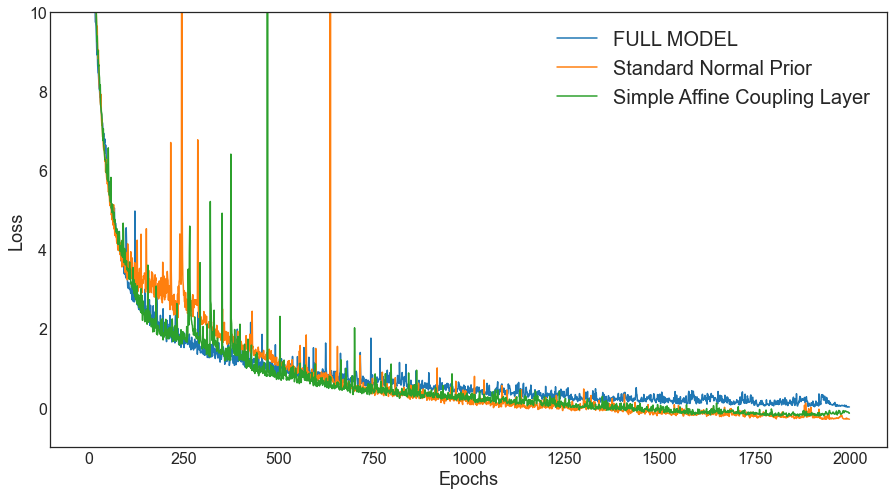}
        \caption{
            Conditioning Ablations Learning Curves Comparison.
        }
        \label{fig:conditioning-ablations-learning-curve}
    \end{figure}

    Additionally,
        figure \ref{fig:conditioning-ablations-learning-curve} compares the loss learning curve of each ablation model.
    The full model presents a more stable curve and higher end-loss, 
        indicating a more regularized model with less over-fitting.  
    
%% file: supplementary_material/experiments/figures/fig-simple-affine-ablation-schemes.tex
\begin{figure}[h]
        \centering
        \begin{minipage}{7cm}
            \includegraphics[height=6cm, width=6cm]
            {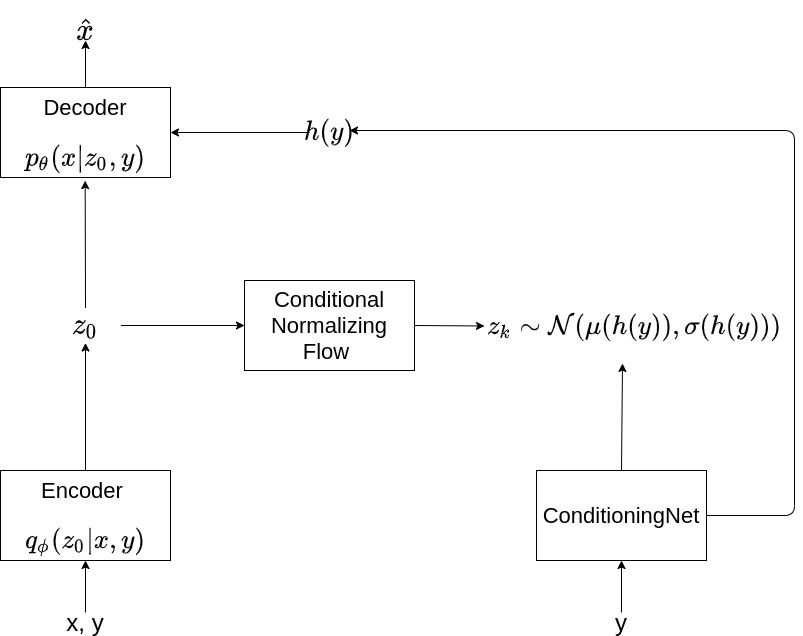}
            \caption*{
                (a) \space Training scheme.
            }
            % \label{fig:cfvae-train-scheme}
        \end{minipage}
        \qquad
        \begin{minipage}{7cm}
            \includegraphics[height=6cm, width=6cm]
            {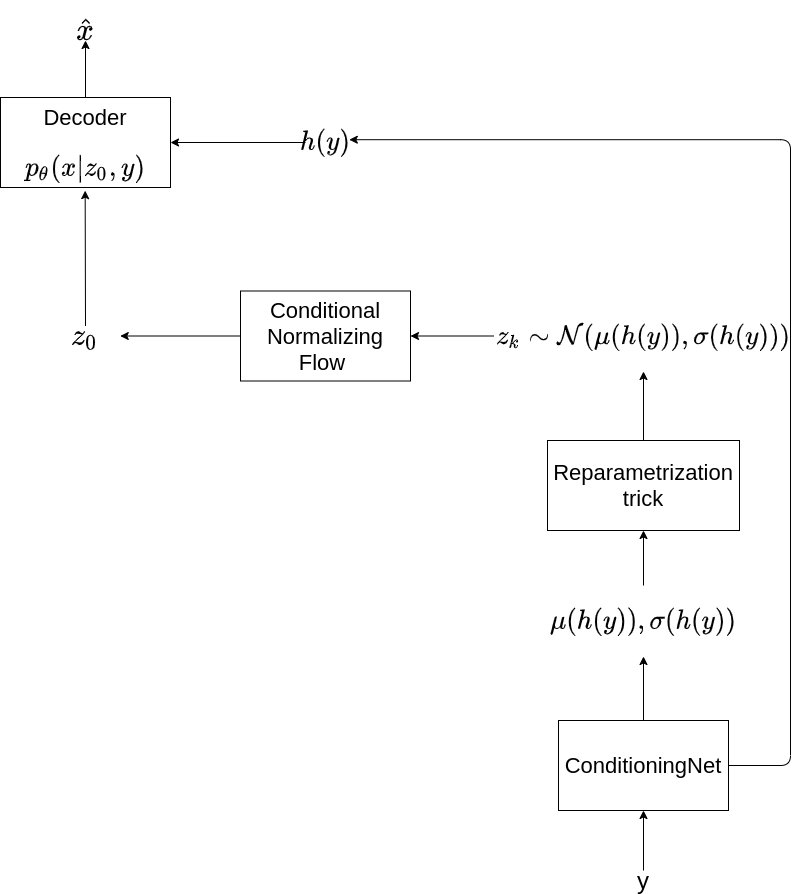}
            \caption*{
                (b) \space Inference scheme.
            }
            % \label{fig:cfvae-test-scheme}
        \end{minipage}
        \caption{
                 Simple Affine Layer ablation train and inference schemes.
            }
            \label{fig:simple-affine-train-and-inference-schemes}
    \end{figure}%

%% file: supplementary_material/experiments/figures/fig-standard-normal-ablation-scheme.tex
\begin{figure}[h]
        \centering
        \begin{minipage}{7cm}
            \includegraphics[height=6cm, width=6cm]
            {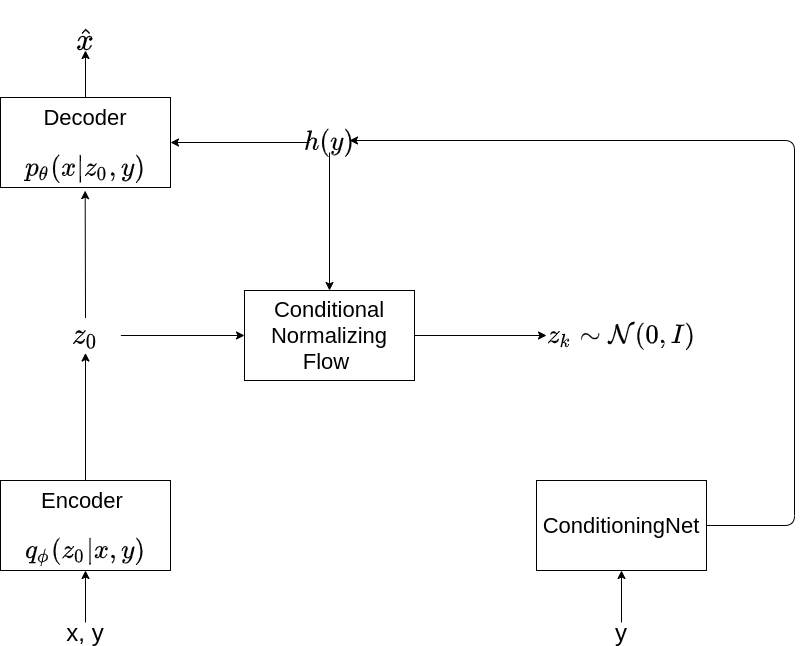}
            \caption*{
                (a) \space Training scheme.
            }
            % \label{fig:cfvae-train-scheme}
        \end{minipage}
        \qquad
        \begin{minipage}{7cm}
            \includegraphics[height=6cm, width=6cm]
            {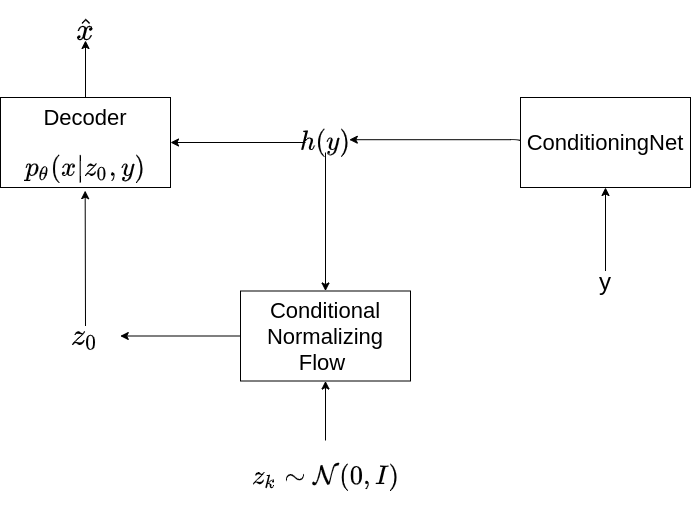}
            \caption*{
                (b) \space Inference scheme.
            }
            % \label{fig:cfvae-test-scheme}
        \end{minipage}
        \caption{
                Standard Normal ablation train and inference schemes.
            }
            \label{fig:standard-normal-train-and-inference-schemes}
    \end{figure}%

%% file: supplementary_material/experiments/figures/fig-full-ablation-scheme.tex
\begin{figure}[h]
        \centering
        \begin{minipage}{7cm}
            \includegraphics[height=6cm, width=6cm]
            {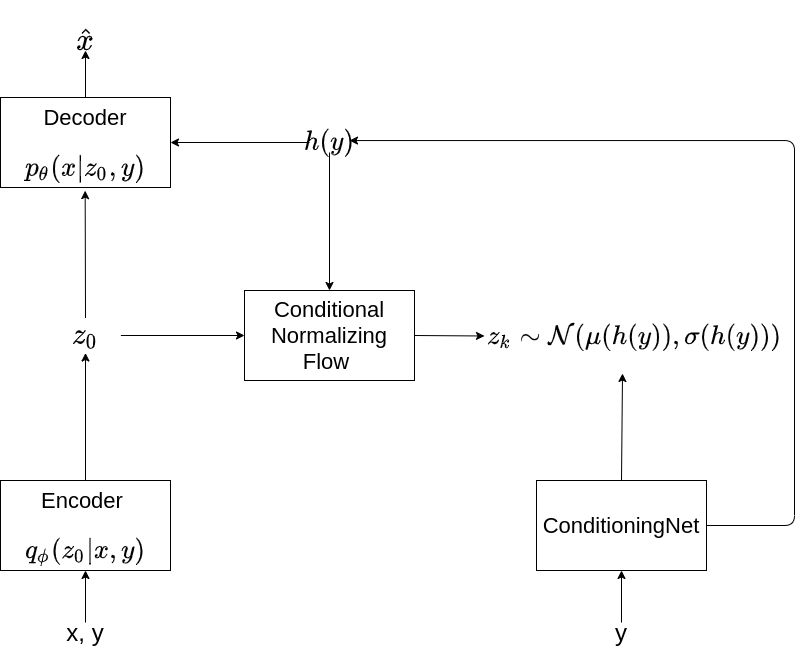}
            \caption*{
                (a) \space Training scheme.
            }
            % \label{fig:cfvae-train-scheme}
        \end{minipage}
        \qquad
        \begin{minipage}{7cm}
            \includegraphics[height=6cm, width=6cm]
            {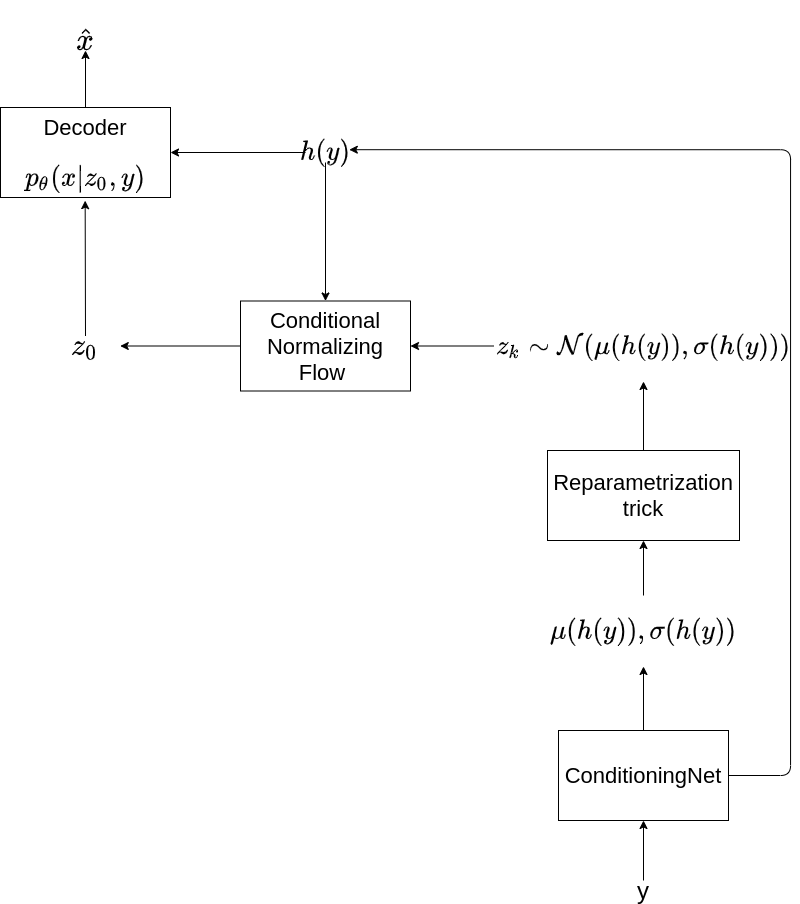}
            \caption*{
                (b) \space Inference scheme.
            }
            % \label{fig:cfvae-test-scheme}
        \end{minipage}
        \caption{
                Full {\ModelName} model train and inference schemes.
            }
            \label{fig:full-model-train-and-inference-schemes}
    \end{figure}%

%% file: supplementary_material/experiments/tables/conditioning_ablations_table.tex
% \newcolumntype{R}{>{\raggedleft\arraybackslash}X}%
% \newcolumntype{L}[1]{>{\raggedright\arraybackslash}X}%
\begin{table*}[ht]
    \caption{
        Per-machine performance comparison between {\ModelName}'s ablation variants.
    }
    \label{tab:cnf-ablations-per-machine}
    \begin{center}
    \begin{small}
    \begin{sc}
    \begin{tabular}{l rr|rr|rr}
        \toprule
        
        {} & 
        \multicolumn{2}{c}{Simple Affine} & 
        \multicolumn{2}{c}{Standard Normal} & 
        \multicolumn{2}{c}{Full Model} 
        
        \\
        
        {} &           
        \multicolumn{1}{c}{SAE} &       
        \multicolumn{1}{c}{NDE} &
        
        \multicolumn{1}{c}{SAE} &       
        \multicolumn{1}{c}{NDE} &        
        
        \multicolumn{1}{c}{SAE} &       
        \multicolumn{1}{c}{NDE}
        
        \\
        \midrule
        
        PI    &      
            0.034 &  
            0.083 &        
            0.149 &  
            0.242 &   
            \textbf{0.018} &  
            \textbf{0.042} 
        \\
        
        PII   &      
            0.036 &  
            0.071 &        
            0.154 & 
            0.224 &   
            \textbf{0.020} &  
            \textbf{0.044} 
        \\
        
        DPCI  &      
            \textbf{0.030} &  
            \textbf{0.046} &        
            0.535 &  
            0.695 &   
            0.035 &  
            0.052 
        \\
        
        DPCII &      
            \textbf{0.029} &  
            \textbf{0.048} &        
            0.542 &  
            0.701 &   
            0.036 &  
            0.055 
        \\
        
        EFI   &      
            \textbf{0.046} &  
            0.153 &        
            0.759 &  
            2.863 &   
            0.047 &  
            \textbf{0.131}
        \\
        
        EFII  &      
            \textbf{0.022} &  
            0.047 &        
            0.329 &  
            0.574 &   
            0.027 &  
            \textbf{0.046} 
        \\
        % \bottomrule
        \midrule
        
        TOTAL &      
            0.200 &  
            0.452 &        
            2.470 &  
            5.301 &   
            \textbf{0.185} &  
            \textbf{0.373} \\
        \bottomrule
    \end{tabular}
    \end{sc}
    \end{small}
    \end{center}
    \vskip -0.15in
\end{table*}

%% file: supplementary_material/experiments/step-flows-ablation.tex
\subsection*{Step-Flows}  \label{sec:step-flow-ablations}
    Section \ref{sec:experiments} of the paper presents ablation studies in the number of step-flow blocks used in the {\ModelName}'s CNF component.
    Five models were trained in the experiment,
        each with $2$, $4$, $8$, $16$, and $32$ 
        step-flow blocks in the CNF component.
        
    Table \ref{tab:cnf-nde-ablations} compares the achieved NDE, 
        per machine,
        by the five trained models.
    \input{supplementary_material/experiments/tables/nde-step-flow_ablations_table}

    Table \ref{tab:cnf-sae-ablations} compares the achieved SAE, 
        per machine, 
        by the five trained models.
    \input{supplementary_material/experiments/tables/sae-step-flow_ablations_table}
    
\begin{figure}[h]
        \centering
        \includegraphics[width=0.9\textwidth]
        {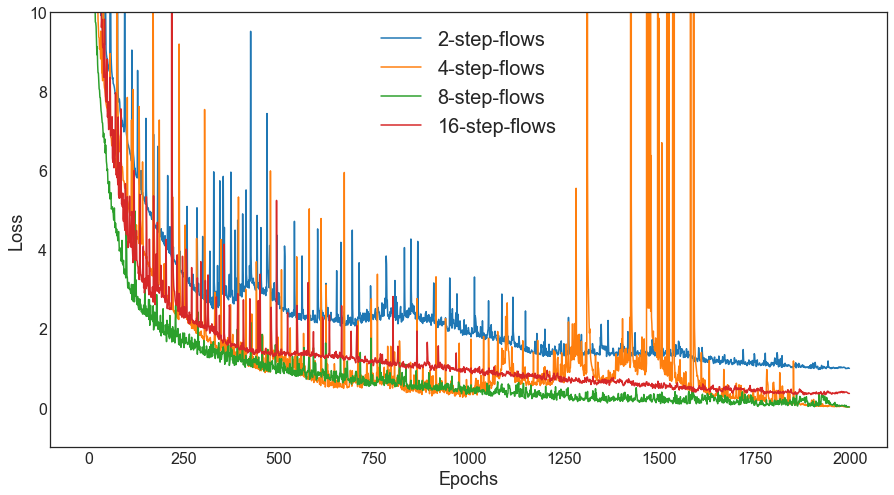}
        \caption{
            Step-Flows Ablations Learning Curves Comparison.
        }
        \label{fig:stepflows-ablations-learning-curve}
    \end{figure}

    Additionally,
        figure \ref{fig:stepflows-ablations-learning-curve} compares the loss learning curve of each ablation model.
    The models with 8 and 16 step-flow blocks have more stable learning curves than the others.
    However,
        the model with 4 step-flow blocks has the lowest end-loss, 
        which is very close to the 8 blocks model's end-loss.
    We believe that it indicates that the number of step-flow blocks influences in both
        the CNF representational capability
        and in the model regularization.
    Thus, 
        the greater the number of step-flow blocks, the greater the CNF representational capability becomes, 
        and the more regularized the model.
    This could explain the reason why model with 16 step-flow blocks
    presented an end-loss greater than the model with 4 step-flow blocks,
    despite having a more stable learning curve.

%% file: supplementary_material/experiments/tables/nde-step-flow_ablations_table.tex
% \newcolumntype{R}{>{\raggedleft\arraybackslash}X}%
% \newcolumntype{L}{>{\raggedright\arraybackslash}X}%
\begin{table}[!htb]
    \caption{
        NDE comparison between {\ModelName} variants with 2, 4, 8, 16, and 32 step-flow blocks in the CNF component.
    }
    \label{tab:cnf-nde-ablations}
    \begin{center}
    \begin{small}
    \begin{sc}
    \begin{tabular}{l rrrrr}
        \toprule
        {} &
        \multicolumn{5}{c}{NDE} \\
        {} &         
        \multicolumn{1}{c}{2} &         
        \multicolumn{1}{c}{4} &         
        \multicolumn{1}{c}{\textbf{8}} &        
        \multicolumn{1}{c}{16} &        
        \multicolumn{1}{c}{32} 
        \\
        \midrule
        
        PI    &  
            0.070 &  0.094 &  \textit{0.042} &  0.051 &  \textbf{0.039} \\
            
        PII   &  
            0.085 &  0.064 &  \textbf{\textit{0.044}} &  0.046 &  0.046 \\
            
        DPCI  &  
            0.337 &  0.073 &  \textbf{\textit{0.052}} &  0.107 &  0.513 \\
            
        DPCII &  
            0.412 &  0.076 &  \textbf{\textit{0.055}} &  0.118 &  0.529 \\
            
        EFI   &  
            1.192 &  0.220 &  \textbf{\textit{0.131}} &  0.272 &  2.838 \\
            
        EFII  &  
            0.218 &  0.083 &  \textbf{\textit{0.046}} &  0.066 &  0.228 \\
        
        % \bottomrule
        \midrule
        TOTAL &  
            2.317 &  0.611 &  \textbf{\textit{0.373}} &  0.663 &  4.196 \\
        
        \bottomrule
    \end{tabular}
    \end{sc}
    \end{small}
    \end{center}
    \vskip -0.15in
\end{table}

%% file: supplementary_material/experiments/tables/sae-step-flow_ablations_table.tex
\begin{table}[!htb]
    \caption{
        SAE comparison between {\ModelName} variants with 2, 4, 8, 16, and 32 step-flow blocks in the CNF component.
    }
    \label{tab:cnf-sae-ablations}
    \begin{center}
    \begin{small}
    \begin{sc}
    \begin{tabular}{l rrrrr}
        \toprule
        {} &
        \multicolumn{5}{c}{SAE} \\
        {} &         
        \multicolumn{1}{c}{2} &         
        \multicolumn{1}{c}{4} &         
        \multicolumn{1}{c}{\textbf{8}} &        
        \multicolumn{1}{c}{16} &        
        \multicolumn{1}{c}{32} 
        \\
        \midrule

        PI    &  
            0.036 &  0.037 &  \textbf{\textit{0.018}} &  0.035 &  0.036 \\
            
        PII   &  
            0.043 &  0.031 &  \textbf{\textit{0.020}} &  0.033 &  0.036 \\
            
        DPCI  &  
            0.280 &  0.043 &  \textbf{\textit{0.035}} &  0.088 &  0.355 \\
            
        DPCII &  
            0.272 &  0.045 &  \textbf{\textit{0.035}} &  0.093 &  0.344 \\
            
        EFI   &  
            0.388 &  0.062 &  \textbf{\textit{0.047}} &  0.118 &  0.631 \\
            
        EFII  &  
            0.168 &  0.040 &  \textbf{\textit{0.027}} &  0.062 &  0.194 \\
        
        \midrule
        
        TOTAL &  
            1.189 &  0.261 &  \textbf{\textit{0.185}} &  0.431 &  1.598 \\
        
        \bottomrule
    \end{tabular}
    \end{sc}
    \end{small}
    \end{center}
    \vskip -0.15in

\end{table}

%% file: supplementary_material/experiments/Visual-Comparison-Plots.tex
    This section presents plots 
        for the visual comparison between 
        the model estimate 
        and the actual active power, 
        the ground truth, 
        for six of the eight devices in the data set.
    Here all presented estimations are taken from the full {\ModelName}, 
        with 8 step-flows blocks 
        and trained with $80\%$ of the dataset.
    Comparisons are made in the dataset test set.
    Moreover, 
        all plots present the actual active power (ground truth), 
        the mean of 10 samples taken from the model,
        and the estimated confidence interval computed at the $95\%$ confidence level. 
    
    \subsection*{Palletizer I - PI}
\begin{figure}[!htb]
                \centering
                \includegraphics[width=\textwidth]
                {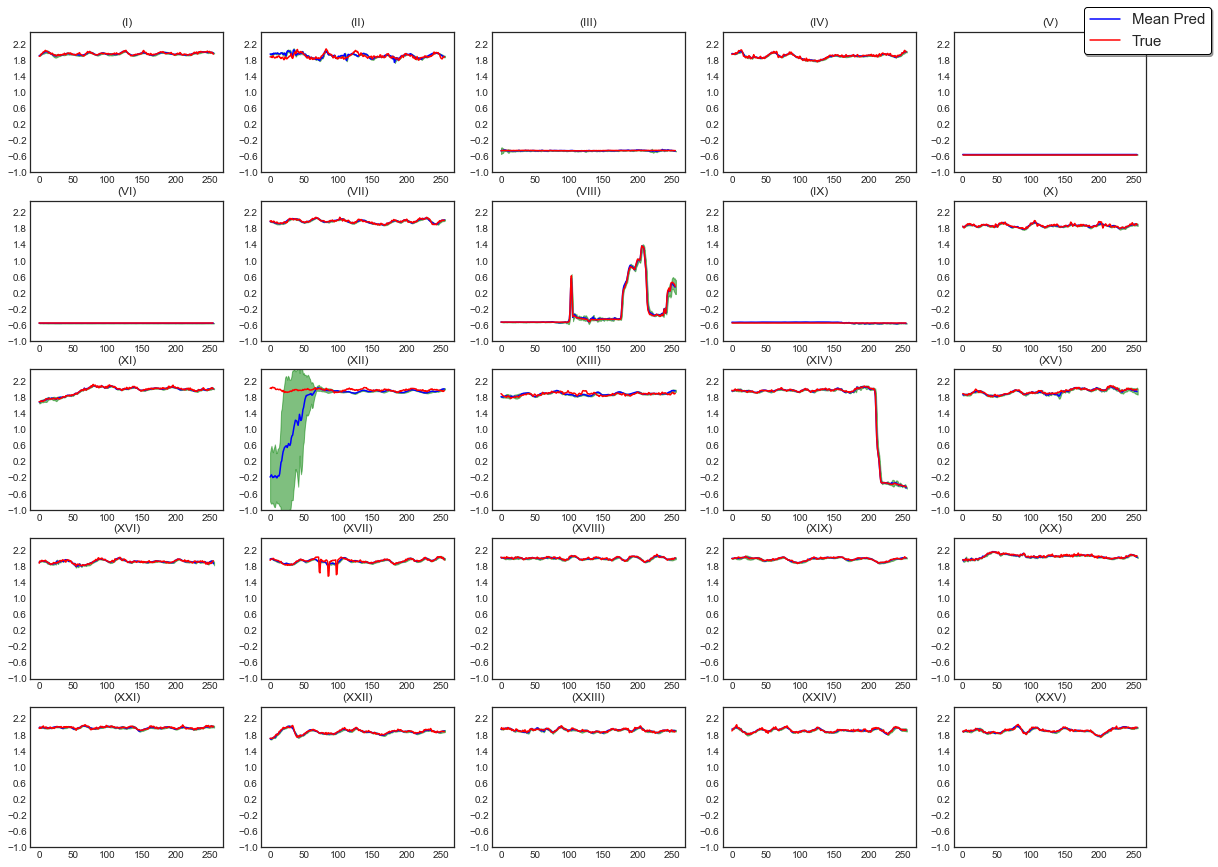}
                \caption{
                    Visual comparison between the model estimates and the Palletizer machine I (PI) ground truth.
                }
                \label{fig:PI-25-comparisons}
            \end{figure}

        Figure \ref{fig:PI-25-comparisons} presents 25 {\ModelName} estimates compared to the PI machine's ground thruth.
        
\begin{figure}[!htb]
                \centering
                \includegraphics[width=\textwidth]
                {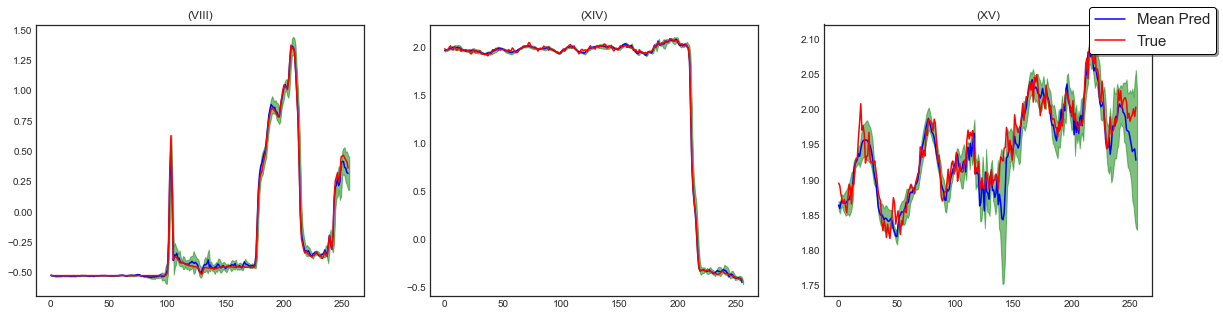}
                \caption{
                    PI re-scaled plots.
                }
                \label{fig:PI-comparisons}
            \end{figure}

        Additionally, 
            figure \ref{fig:PI-comparisons} presents re-scaled plots for 3 of the 25 comparisons to observe the signal in a more detailed view and illustrate the inferred confidence interval (the green shaded area).
        
    \subsection*{Palletizer II - PII}
\begin{figure}[!htb]
                \centering
                \includegraphics[width=\textwidth]
                {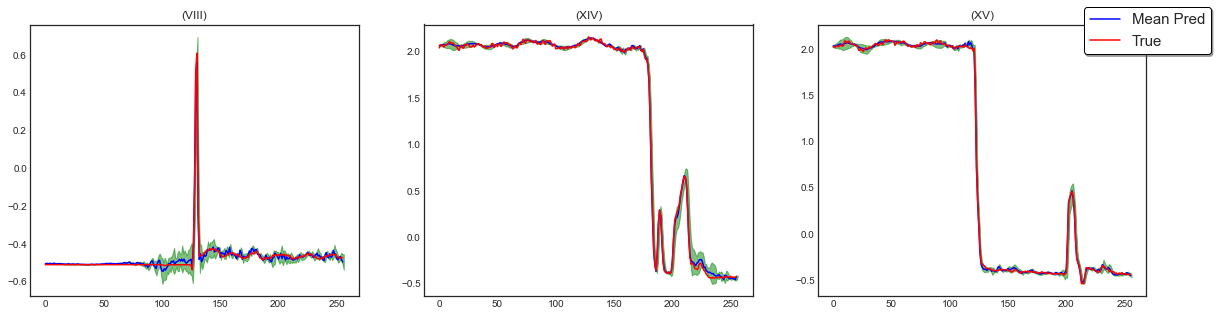}
                \caption{
                    PII re-scaled plots.
                }
                \label{fig:PII-comparisons}
            \end{figure}

        Figure \ref{fig:PII-comparisons} presents 25 {\ModelName} estimates compared to the PII machine's ground truth.
        \begin{figure}[!htb]
                \centering
                \includegraphics[width=\textwidth]
                {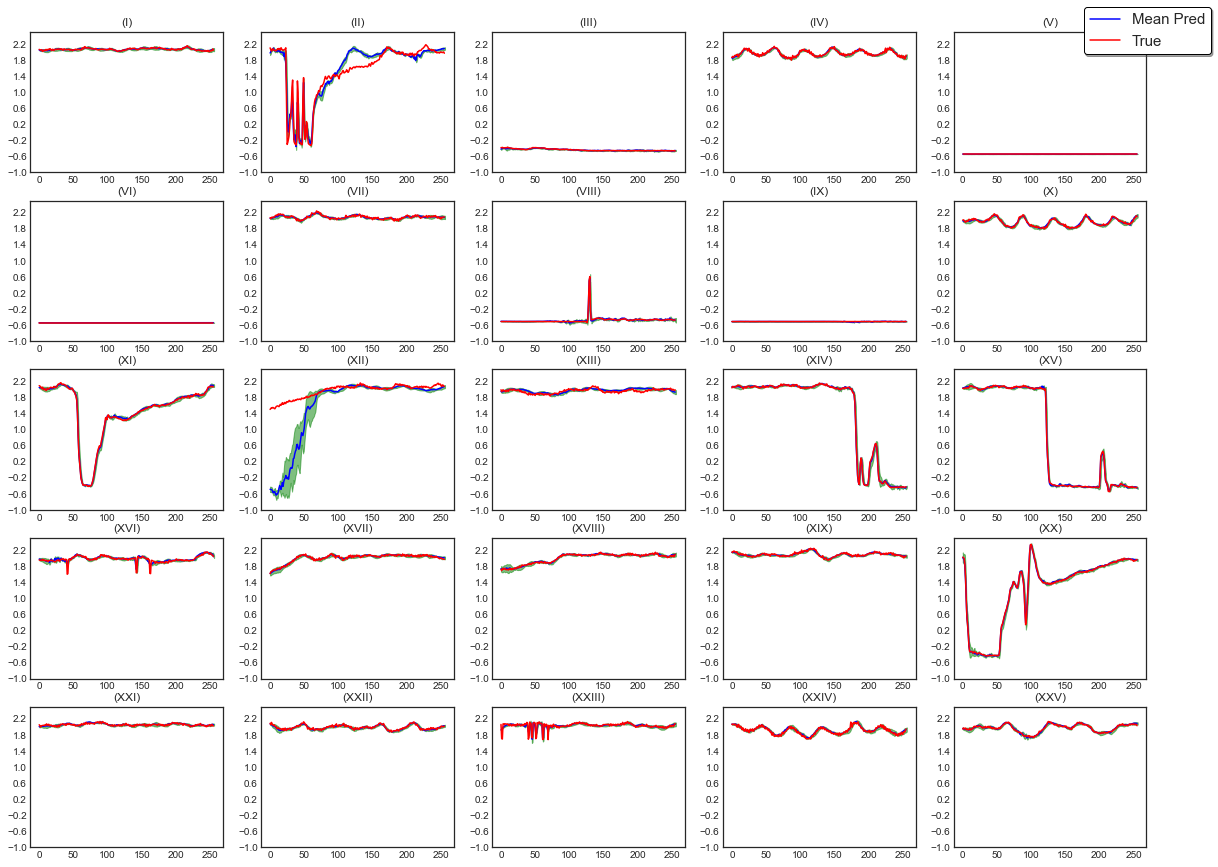}
                \caption{
                    Visual comparison between the model estimates and the Palletizer machine II (PII) ground truth.
                }
                \label{fig:PII-25-comparisons}
            \end{figure}%
        
        Additionally, 
            figure \ref{fig:PII-comparisons} presents re-scaled plots for 3 of the 25 comparisons 
            to observe the signal in a more detailed view and illustrate the inferred confidence interval 
            (the green shaded area).
    
    \subsection*{Pouble-Pole Contactor I - DPCI}
\begin{figure}[!htb]
                \centering
                \includegraphics[width=\textwidth]
                {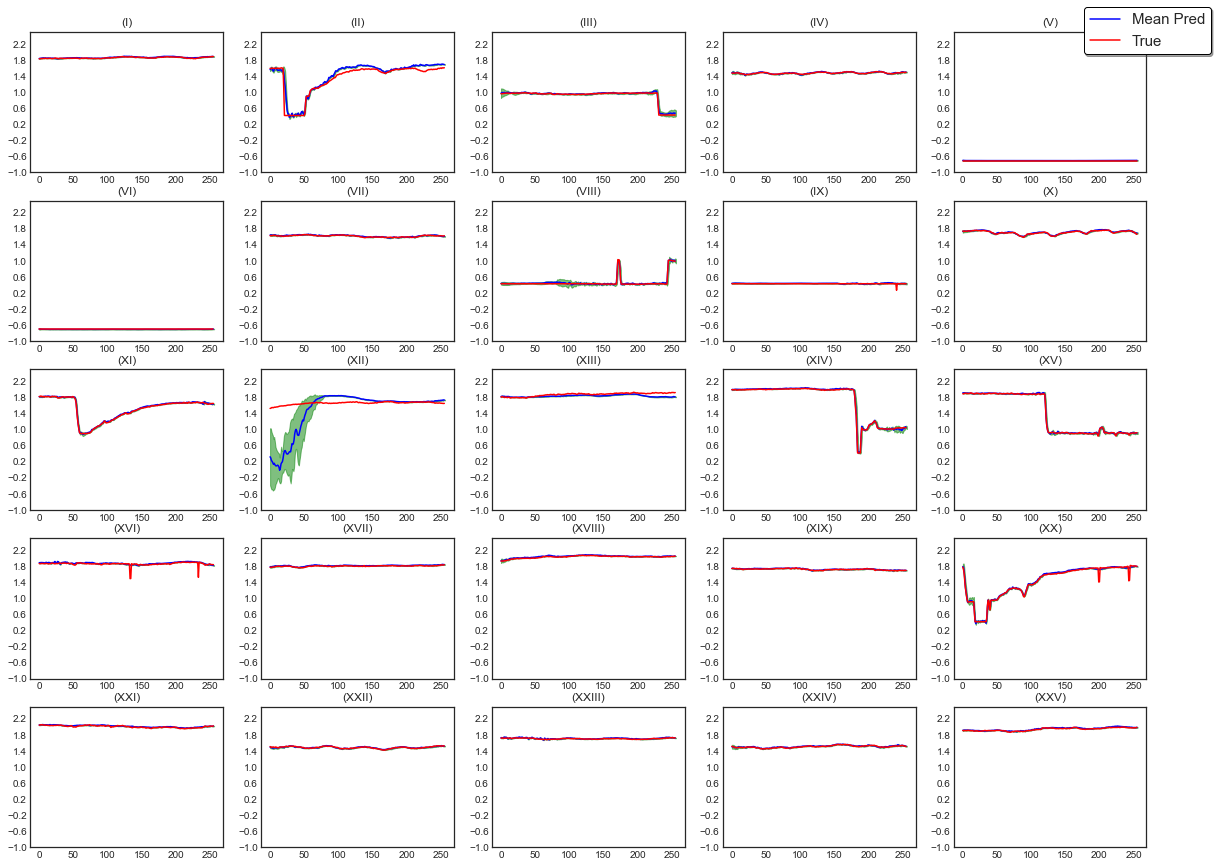}
                \caption{
                    Visual comparison between the model estimates and the ground truth for the Double-Pole Contactor machine I (DPCI).
                }
                \label{fig:DPCI-25-comparisons}
            \end{figure}

        Figure \ref{fig:DPCI-25-comparisons} presents 25 {\ModelName} estimates compared to the DPCI machine's ground truth.
        
\begin{figure}[!htb]
                \centering
                \includegraphics[width=\textwidth]
                {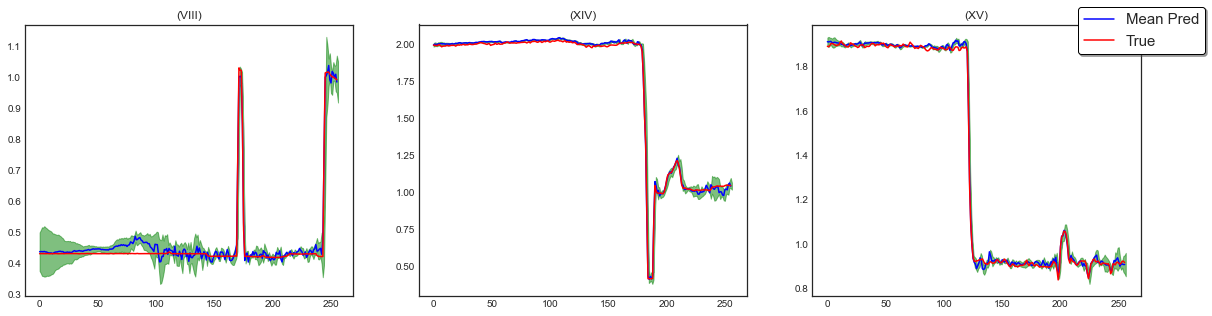}
                \caption{
                    DPCI re-scaled plots.
                }
                \label{fig:DPCI-comparisons}
            \end{figure}

        Additionally, 
            figure \ref{fig:DPCI-comparisons} presents re-scaled plots for 3 of the 25 comparisons 
            to observe the signal in a more detailed view and illustrate the inferred confidence interval 
            (the green shaded area).
    
    \subsection*{Double-Pole Contactorr II - DPCII}
\begin{figure}[!htb]
                \centering
                \includegraphics[width=\textwidth]
                {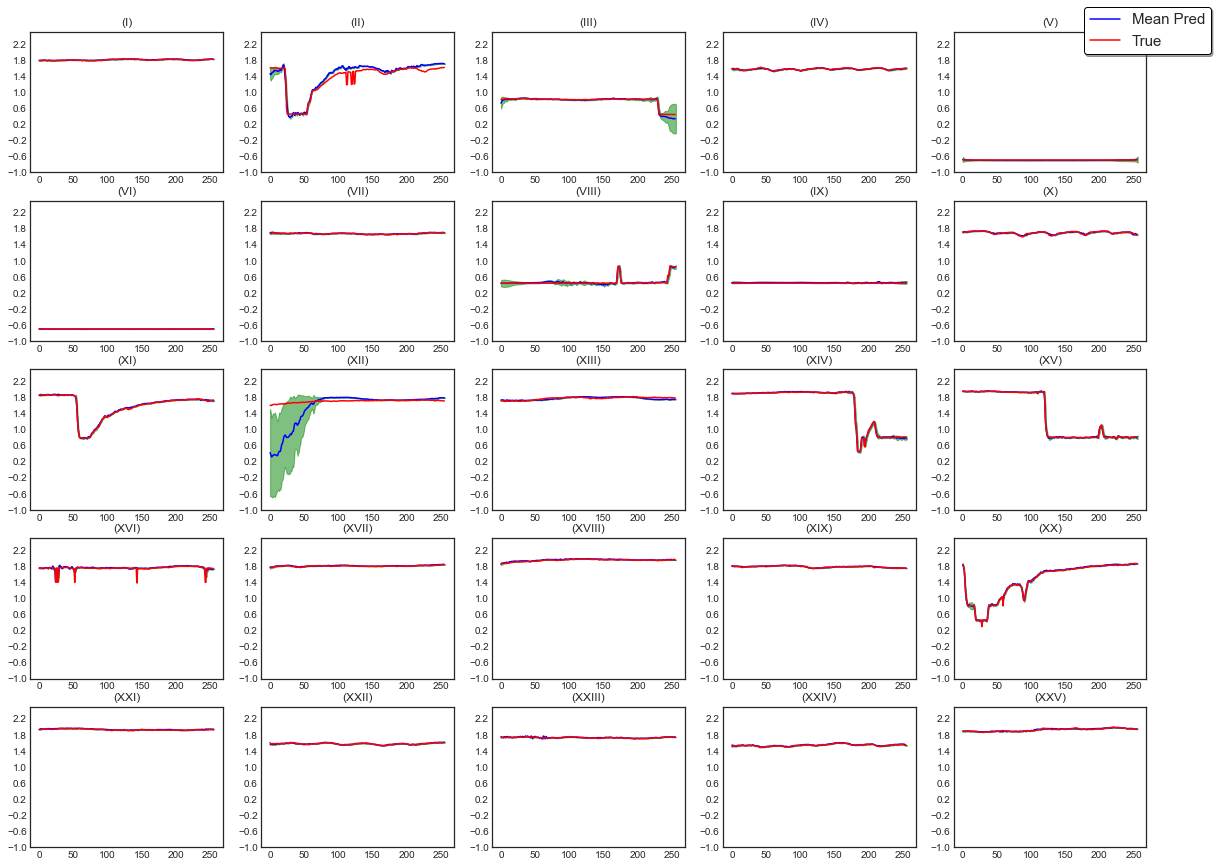}
                \caption{
                    Visual comparison between the model estimates and the Double-Pole Contactor machine II (DPCII) ground truth.
                }
                \label{fig:DPCII-25-comparisons}
            \end{figure}

        Figure \ref{fig:DPCII-25-comparisons} presents 25 {\ModelName} estimates compared to the DPCII machine's ground truth.
        
\begin{figure}[!htb]
                \centering
                \includegraphics[width=\textwidth]
                {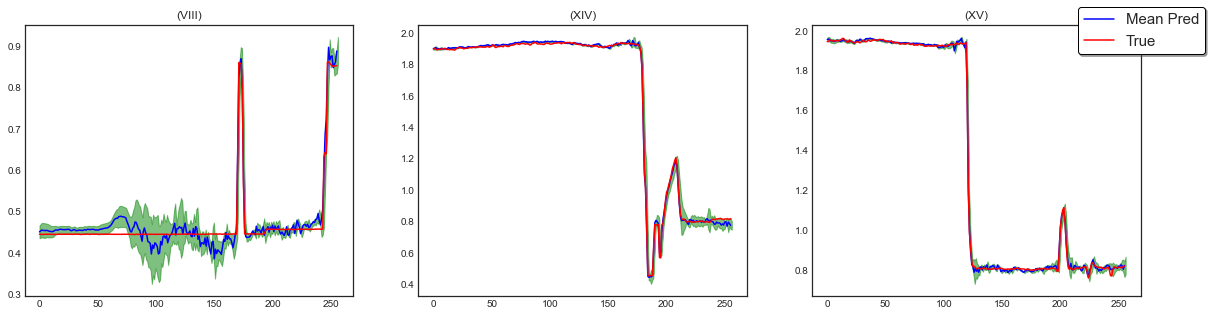}
                \caption{
                    DPCII re-scaled plots.
                }
                \label{fig:DPCII-comparisons}
            \end{figure}

        Additionally, 
            figure \ref{fig:DPCII-comparisons} presents re-scaled plots for 3 of the 25 comparisons 
            to observe the signal in a more detailed view and illustrate the inferred confidence interval 
            (the green shaded area).
            
    \subsection*{Exhaust Fan I - EFI}
\begin{figure}[!htb]
                \centering
                \includegraphics[width=\textwidth]
                {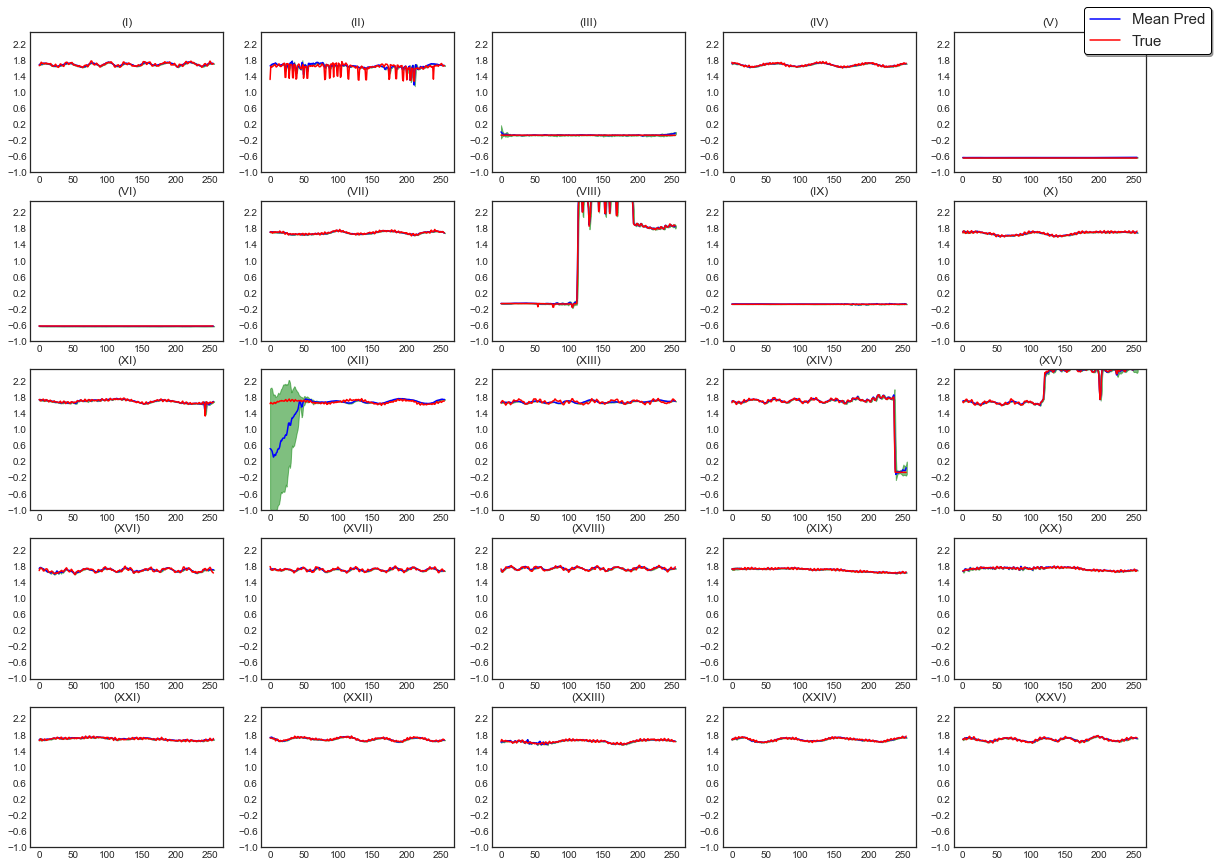}
                \caption{
                    Visual comparison between the model estimates and the Exhaust Fan machine I (EFI) ground truth.
                }
                \label{fig:EFI-25-comparisons}
            \end{figure}

        Figure \ref{fig:EFI-25-comparisons} presents 25 {\ModelName} estimates compared to the EFI machine's ground truth.
        
\begin{figure}[!htb]
                \centering
                \includegraphics[width=\textwidth]
                {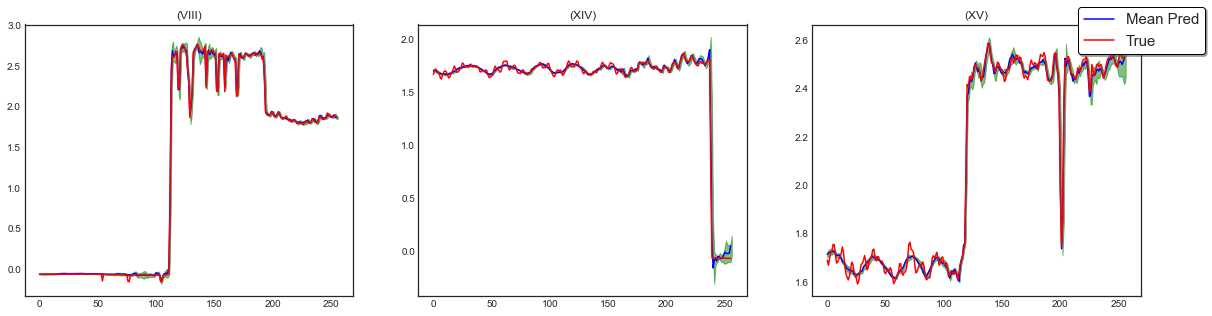}
                \caption{
                    EFI re-scaled plots.
                }
                \label{fig:EFI-comparisons}
            \end{figure}

        Additionally, 
            figure \ref{fig:EFI-comparisons} presents re-scaled plots for 3 of the 25 comparisons 
            to observe the signal in a more detailed view and illustrate the inferred confidence interval 
            (the green shaded area).
    
    \subsection*{Exhaust Fan II - EFII}
\begin{figure}[!htb]
                \centering
                \includegraphics[width=\textwidth]
                {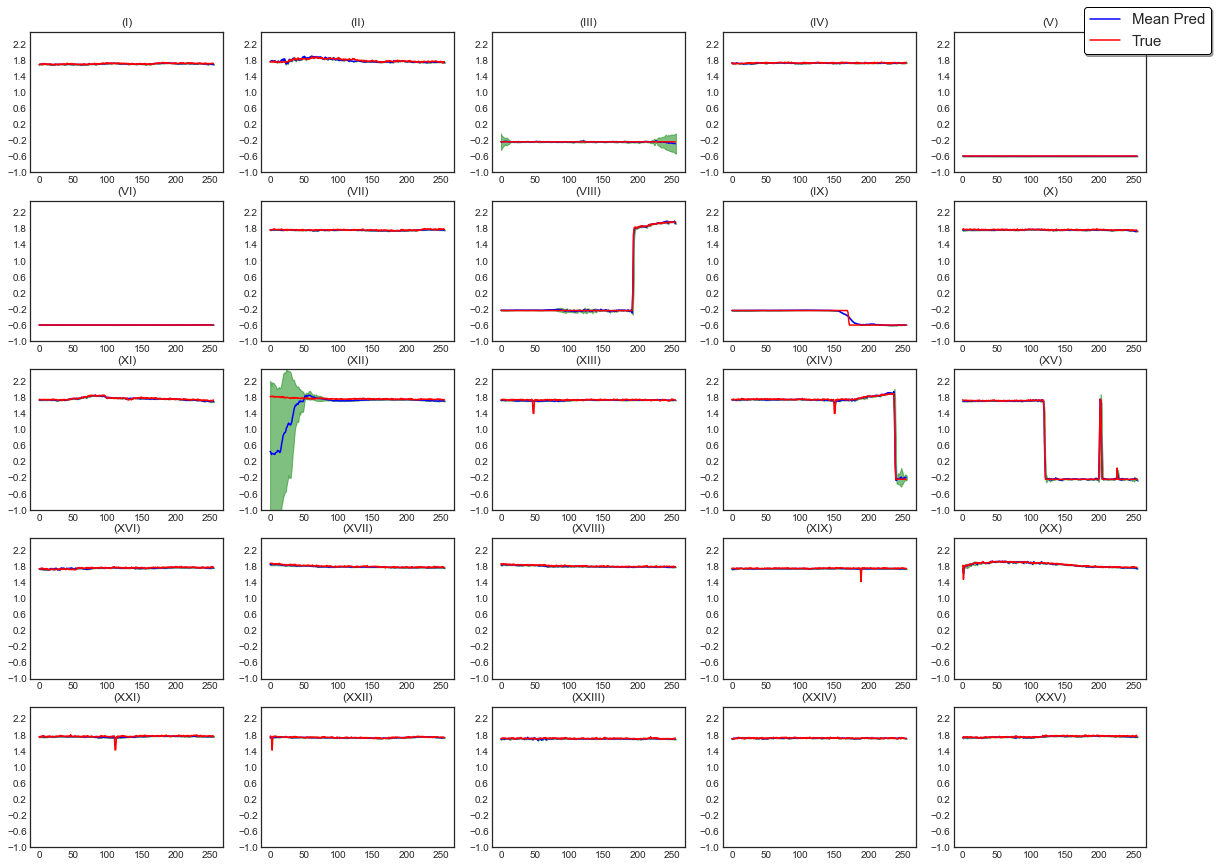}
                \caption{
                    Visual comparison between the model estimates and the ground truth for the Exhaust Fan machine II (EFII).
                }
                \label{fig:EFII-25-comparisons}
            \end{figure}

        Figure \ref{fig:EFII-25-comparisons} presents 25 {\ModelName} estimates compared to the EFII machine's ground truth.
        
\begin{figure}[!htb]
                \centering
                \includegraphics[width=\textwidth]
                {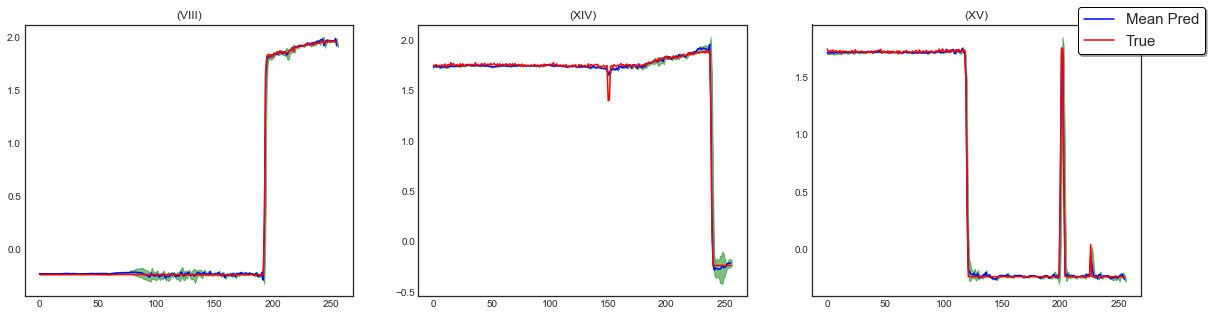}
                \caption{
                    EFII re-scaled plots.
                }
                \label{fig:EFII-comparisons}
            \end{figure}

        Additionally, 
            figure \ref{fig:EFII-comparisons} presents re-scaled plots for 3 of the 25 comparisons 
            to observe the signal in a more detailed view and illustrate the inferred confidence interval 
            (the green shaded area).

%% file: main_ieee.bbl
\begin{thebibliography}{10}
\providecommand{\url}[1]{#1}
\csname url@samestyle\endcsname
\providecommand{\newblock}{\relax}
\providecommand{\bibinfo}[2]{#2}
\providecommand{\BIBentrySTDinterwordspacing}{\spaceskip=0pt\relax}
\providecommand{\BIBentryALTinterwordstretchfactor}{4}
\providecommand{\BIBentryALTinterwordspacing}{\spaceskip=\fontdimen2\font plus
\BIBentryALTinterwordstretchfactor\fontdimen3\font minus
  \fontdimen4\font\relax}
\providecommand{\BIBforeignlanguage}[2]{{%
\expandafter\ifx\csname l@#1\endcsname\relax
\typeout{** WARNING: IEEEtran.bst: No hyphenation pattern has been}%
\typeout{** loaded for the language `#1'. Using the pattern for}%
\typeout{** the default language instead.}%
\else
\language=\csname l@#1\endcsname
\fi
#2}}
\providecommand{\BIBdecl}{\relax}
\BIBdecl

\bibitem{Kelly2015NeuralND}
J.~Kelly and W.~J. Knottenbelt, ``Neural nilm: Deep neural networks applied to
  energy disaggregation,'' \emph{ArXiv}, vol. abs/1507.06594, 2015.

\bibitem{Martins2018ApplicationOA}
P.~B.~M. Martins, J.~G. R.~C. Gomes, V.~B. Nascimento, and A.~R. de~Freitas,
  ``Application of a deep learning generative model to load disaggregation for
  industrial machinery power consumption monitoring,'' \emph{2018 IEEE
  International Conference on Communications, Control, and Computing
  Technologies for Smart Grids}, pp. 1--6, 2018.

\bibitem{Roos1994UsingNN}
J.~G. Roos, I.~E. Lane, E.~C. Botha, and G.~P. Hancke, ``Using neural networks
  for non-intrusive monitoring of industrial electrical loads,''
  \emph{Conference Proceedings. 10th Anniversary. IMTC/94. Advanced
  Technologies in I \& M. 1994 IEEE Instrumentation and Measurement Technolgy
  Conference (Cat. No.94CH3424-9)}, pp. 1115--1118 vol.3, 1994.

\bibitem{Yang&Chang2007}
H.~{Yang}, H.~{Chang}, and C.~{Lin}, ``Design a neural network for features
  selection in non-intrusive monitoring of industrial electrical loads,'' in
  \emph{2007 11th International Conference on Computer Supported Cooperative
  Work in Design}, 2007, pp. 1022--1027.

\bibitem{Lin&Tsai2010}
Y.~{Lin} and M.~{Tsai}, ``A novel feature extraction method for the development
  of nonintrusive load monitoring system based on bp-ann,'' in \emph{2010
  International Symposium on Computer, Communication, Control and Automation
  (3CA)}, vol.~2, 2010, pp. 215--218.

\bibitem{Ruzzelli&Nicolas&Schoofs2010}
A.~G. {Ruzzelli}, C.~{Nicolas}, A.~{Schoofs}, and G.~M.~P. {O'Hare},
  ``Real-time recognition and profiling of appliances through a single
  electricity sensor,'' in \emph{2010 7th Annual IEEE Communications Society
  Conference on Sensor, Mesh and Ad Hoc Communications and Networks}, 2010, pp.
  1--9.

\bibitem{Chang&Chien&Lin&Chen2011}
H.~{Chang}, P.~{Chien}, L.~{Lin}, and N.~{Chen}, ``Feature extraction of
  non-intrusive load-monitoring system using genetic algorithm in smart
  meters,'' in \emph{2011 IEEE 8th International Conference on e-Business
  Engineering}, 2011, pp. 299--304.

\bibitem{Hart1992NonintrusiveAL}
G.~Hart, ``Nonintrusive appliance load monitoring,'' \emph{Proceedings of the
  IEEE}, vol.~80, no.~12, pp. 1870--1891, 1992.

\bibitem{Kingma2013AutoEncodingVB}
D.~P. Kingma and M.~Welling, ``Auto-encoding variational bayes,'' \emph{CoRR},
  vol. abs/1312.6114, 2013.

\bibitem{Sohn2015LearningSO}
K.~Sohn, H.~Lee, and X.~Yan, ``Learning structured output representation using
  deep conditional generative models,'' in \emph{NIPS}, 2015.

\bibitem{Dinh2014NICENI}
L.~Dinh, D.~Krueger, and Y.~Bengio, ``Nice: Non-linear independent components
  estimation,'' \emph{CoRR}, vol. abs/1410.8516, 2014.

\bibitem{Dinh2016DensityEU}
L.~Dinh, J.~Sohl-Dickstein, and S.~Bengio, ``Density estimation using real
  nvp,'' \emph{ArXiv}, vol. abs/1605.08803, 2016.

\bibitem{Kingma2018GlowGF}
D.~P. Kingma and P.~Dhariwal, ``Glow: Generative flow with invertible 1x1
  convolutions,'' in \emph{NeurIPS}, 2018.

\bibitem{Martins2018Dataset}
\BIBentryALTinterwordspacing
P.~B. de~Mello Martins; Vagner Barbosa Nascimento; Antônio Renato de Freitas;
  Pedro Bittencourt~e Silva; Raphael Guimarães Duarte~Pinto, ``Industrial
  machines dataset for electrical load disaggregation,'' 2018. [Online].
  Available: \url{http://dx.doi.org/10.21227/cg5v-dk02}
\BIBentrySTDinterwordspacing

\bibitem{Zeifman2011NonintrusiveAL}
M.~Zeifman and K.~Roth, ``Nonintrusive appliance load monitoring: Review and
  outlook,'' \emph{IEEE Transactions on Consumer Electronics}, vol.~57, 2011.

\bibitem{Armel2013IsDT}
K.~C. Armel, A.~Gupta, G.~Shrimali, and A.~Albert, ``Is disaggregation the holy
  grail of energy efficiency? the case of electricity,'' \emph{Energy Policy},
  vol.~52, pp. 213--234, 2013.

\bibitem{UK-DALE}
J.~Kelly and W.~Knottenbelt, ``The {UK-DALE} dataset, domestic appliance-level
  electricity demand and whole-house demand from five {UK} homes,''
  \emph{Scientific Data}, vol.~2, no. 150007, 2015.

\bibitem{Bonfigli2018DenoisingAF}
R.~Bonfigli, A.~Felicetti, E.~Principi, M.~Fagiani, S.~Squartini, and
  F.~Piazza, ``Denoising autoencoders for non-intrusive load monitoring:
  Improvements and comparative evaluation,'' \emph{Energy and Buildings}, vol.
  158, pp. 1461--1474, 2018.

\bibitem{Kolter2012ApproximateII}
J.~Z. Kolter and T.~Jaakkola, ``Approximate inference in additive factorial
  hmms with application to energy disaggregation,'' in \emph{AISTATS}, 2012.

\bibitem{Zhang2018SequencetopointLW}
C.~Zhang, M.~Zhong, Z.~Wang, N.~Goddard, and C.~Sutton, ``Sequence-to-point
  learning with neural networks for nonintrusive load monitoring,'' in
  \emph{AAAI}, 2018.

\bibitem{Ghahramani1997FactorialHM}
Z.~Ghahramani and M.~I. Jordan, ``Factorial hidden markov models,''
  \emph{Machine Learning}, vol.~29, pp. 245--273, 1997.

\bibitem{Oord2016WaveNetAG}
A.~van~den Oord, S.~Dieleman, H.~Zen, K.~Simonyan, O.~Vinyals, A.~Graves,
  N.~Kalchbrenner, A.~W. Senior, and K.~Kavukcuoglu, ``Wavenet: A generative
  model for raw audio,'' \emph{ArXiv}, vol. abs/1609.03499, 2016.

\bibitem{Jordan1999AnIT}
M.~I. Jordan, Z.~Ghahramani, T.~Jaakkola, and L.~Saul, ``An introduction to
  variational methods for graphical models,'' \emph{Machine Learning}, vol.~37,
  pp. 183--233, 1999.

\bibitem{Walker2016AnUF}
J.~Walker, C.~Doersch, A.~Gupta, and M.~Hebert, ``An uncertain future:
  Forecasting from static images using variational autoencoders,'' in
  \emph{ECCV}, 2016.

\bibitem{Doersch2016TutorialOV}
C.~Doersch, ``Tutorial on variational autoencoders,'' \emph{ArXiv}, vol.
  abs/1606.05908, 2016.

\bibitem{Ardizzone2019GuidedIG}
L.~Ardizzone, C.~L{\"u}th, J.~Kruse, C.~Rother, and U.~K{\"o}the, ``Guided
  image generation with conditional invertible neural networks,'' \emph{ArXiv},
  vol. abs/1907.02392, 2019.

\bibitem{Winkler2019LearningLW}
C.~Winkler, D.~Worrall, E.~Hoogeboom, and M.~Welling, ``Learning likelihoods
  with conditional normalizing flows,'' \emph{ArXiv}, vol. abs/1912.00042,
  2019.

\bibitem{Berg2018SylvesterNF}
R.~V. Berg, L.~Hasenclever, J.~Tomczak, and M.~Welling, ``Sylvester normalizing
  flows for variational inference,'' in \emph{UAI}, 2018.

\bibitem{Bhattacharyya2019ConditionalFV}
A.~Bhattacharyya, M.~Hanselmann, M.~Fritz, B.~Schiele, and C.~Straehle,
  ``Conditional flow variational autoencoders for structured sequence
  prediction,'' \emph{ArXiv}, vol. abs/1908.09008, 2019.

\bibitem{Ziegler2019LatentNF}
Z.~M. Ziegler and A.~M. Rush, ``Latent normalizing flows for discrete
  sequences,'' \emph{ArXiv}, vol. abs/1901.10548, 2019.

\bibitem{Nielsen2020SurVAEFS}
\BIBentryALTinterwordspacing
D.~Nielsen, P.~Jaini, E.~Hoogeboom, O.~Winther, and M.~Welling, ``Survae flows:
  Surjections to bridge the gap between vaes and flows,'' in \emph{Advances in
  Neural Information Processing Systems}, H.~Larochelle, M.~Ranzato,
  R.~Hadsell, M.~F. Balcan, and H.~Lin, Eds., vol.~33.\hskip 1em plus 0.5em
  minus 0.4em\relax Curran Associates, Inc., 2020, pp. 12\,685--12\,696.
  [Online]. Available:
  \url{https://proceedings.neurips.cc/paper/2020/file/9578a63fbe545bd82cc5bbe749636af1-Paper.pdf}
\BIBentrySTDinterwordspacing

\bibitem{Yang2019PointFlow3P}
G.~Yang, X.~Huang, Z.~Hao, M.-Y. Liu, S.~Belongie, and B.~Hariharan,
  ``Pointflow: 3d point cloud generation with continuous normalizing flows,''
  in \emph{2019 IEEE/CVF International Conference on Computer Vision (ICCV)},
  2019, pp. 4540--4549.

\bibitem{Ma2019FlowSeqNC}
\BIBentryALTinterwordspacing
X.~Ma, C.~Zhou, X.~Li, G.~Neubig, and E.~Hovy, ``{F}low{S}eq:
  Non-autoregressive conditional sequence generation with generative flow,'' in
  \emph{Proceedings of the 2019 Conference on Empirical Methods in Natural
  Language Processing and the 9th International Joint Conference on Natural
  Language Processing (EMNLP-IJCNLP)}.\hskip 1em plus 0.5em minus 0.4em\relax
  Hong Kong, China: Association for Computational Linguistics, Nov. 2019, pp.
  4282--4292. [Online]. Available:
  \url{https://www.aclweb.org/anthology/D19-1437}
\BIBentrySTDinterwordspacing

\bibitem{Dauphin2017LanguageMW}
Y.~Dauphin, A.~Fan, M.~Auli, and D.~Grangier, ``Language modeling with gated
  convolutional networks,'' in \emph{ICML}, 2017.

\bibitem{zhong2014signal}
\BIBentryALTinterwordspacing
M.~Zhong, N.~Goddard, and C.~Sutton, ``Signal aggregate constraints in additive
  factorial hmms, with application to energy disaggregation,'' in
  \emph{Advances in Neural Information Processing Systems}, Z.~Ghahramani,
  M.~Welling, C.~Cortes, N.~Lawrence, and K.~Q. Weinberger, Eds.,
  vol.~27.\hskip 1em plus 0.5em minus 0.4em\relax Curran Associates, Inc.,
  2014. [Online]. Available:
  \url{https://proceedings.neurips.cc/paper/2014/file/2d1b2a5ff364606ff041650887723470-Paper.pdf}
\BIBentrySTDinterwordspacing

\bibitem{Kingma2015AdamAM}
D.~P. Kingma and J.~Ba, ``Adam: A method for stochastic optimization,''
  \emph{CoRR}, vol. abs/1412.6980, 2015.

\bibitem{CV:Kohavi:1995}
\BIBentryALTinterwordspacing
R.~Kohavi, ``A study of cross-validation and bootstrap for accuracy estimation
  and model selection,'' in \emph{Proceedings of the 14th International Joint
  Conference on Artificial Intelligence - Volume 2}, ser. IJCAI'95.\hskip 1em
  plus 0.5em minus 0.4em\relax San Francisco, CA, USA: Morgan Kaufmann
  Publishers Inc., 1995, pp. 1137--1143. [Online]. Available:
  \url{http://dl.acm.org/citation.cfm?id=1643031.1643047}
\BIBentrySTDinterwordspacing

\bibitem{kolter2011redd}
J.~Z. Kolter and M.~J. Johnson, ``Redd: A public data set for energy
  disaggregation research,'' in \emph{Workshop on data mining applications in
  sustainability (SIGKDD), San Diego, CA}, vol.~25, no. Citeseer, 2011, pp.
  59--62.

\end{thebibliography}
